\documentclass{article}

\usepackage{microtype}
\usepackage{graphicx}
\usepackage{subfigure}
\usepackage{booktabs} 

\usepackage{hyperref}

\usepackage[accepted]{icml2025}

\usepackage{amsmath}
\usepackage{amssymb}
\usepackage{mathtools}
\usepackage{amsthm}

\usepackage[capitalize,noabbrev]{cleveref}

\usepackage{xcolor}         
\usepackage{array}
\usepackage{multirow}
\usepackage{siunitx}
\usepackage{enumitem}
\usepackage[most]{tcolorbox}
\usepackage{dashrule}
\usepackage{colortbl}  
\usepackage{xfp}

\theoremstyle{plain}

\theoremstyle{definition}

\theoremstyle{remark}

\usepackage[textsize=tiny]{todonotes}

\icmltitlerunning{Selective Prompt Anchoring for Code Generation}

\begin{document}

\definecolor{posgreen}{RGB}{85,245,130}
\definecolor{negred}{RGB}{230,110,110}

\setkeys{Gin}{width=\linewidth, clip}

\newcommand{\tz}[1]{\textcolor{red}{#1}}
\newcommand{\red}[1]{\textcolor{red}{#1}}
\newcommand{\fix}{\marginpar{FIX}}
\newcommand{\new}{\marginpar{NEW}}
\newcommand{\tool}{\textsc{Spa}}
\newcommand{\weight}{\omega}
\newcommand{\logits}{F_{\theta, i, x}}
\newcommand{\embedding}{\mathbf{E}_i}
\newcommand{\model}{f_{\theta}}
 \newcommand{\circled}[1]{{\large\textcircled{\footnotesize #1}}}
\definecolor{mygreen}{HTML}{1FA428}
\definecolor{myred}{HTML}{B22222}
\newcommand{\finding}[1]{{#1}}
\newcommand{\smallerfootnote}{%
  \fontsize{8.5pt}{8.4pt}\selectfont
}

\newcommand{\slightmoresmall}{\fontsize{9.3pt}{5pt}\selectfont}
\newcommand{\moresmall}{\fontsize{8.6pt}{2pt}\selectfont}

\newcommand{\moremoresmall}{\fontsize{7.7pt}{2pt}\selectfont}

\newcommand{\moretiny}{\fontsize{4.9pt}{5.5pt}\selectfont}
\newcommand{\specialtiny}{\fontsize{5.7pt}{5.5pt}\selectfont}

\newcommand{\moresmallerfootnote}{%
  \fontsize{7.6pt}{8.4pt}\selectfont
}
\newcommand{\xxx}[1]{\textbf{\textcolor{red}{#1}}}

\renewcommand{\todo}[1]{%
  \mbox{%
    \colorbox{orange!20}{%
      \begin{minipage}{\dimexpr\linewidth-2\fboxsep\relax}
        {\color{orange!80!black}\small\sffamily#1}%
      \end{minipage}%
    }%
  }%
}

\newcommand{\deltacolor}[1]{%
    \ifnum\fpeval{#1 < 0}=1
        \cellcolor{negred!\fpeval{min(max(abs(#1*10), 10), 100)}!white}%
    \else
        \ifnum\fpeval{#1 = 0}=1
        \else
            \cellcolor{posgreen!\fpeval{min(max(#1*8, 10), 100)}!white}%
        \fi
    \fi
}

\twocolumn[
\icmltitle{Selective Prompt Anchoring for Code Generation}




\begin{icmlauthorlist}
\icmlauthor{Yuan Tian}{purdue}
\icmlauthor{Tianyi Zhang}{purdue}
\end{icmlauthorlist}

\icmlaffiliation{purdue}{Department of Computer Science, Purdue University, West Lafayette, IN, USA}

\icmlcorrespondingauthor{Yuan Tian}{tian211@purdue.edu}
\icmlcorrespondingauthor{Tianyi Zhang}{tianyi@purdue.edu}

\icmlkeywords{Machine Learning, ICML}

\vskip 0.3in
]



\printAffiliationsAndNotice{}  

\begin{abstract}
    Recent advances in large language models (LLMs) have transformed software development by automatically generating code from natural language. Yet challenges remain in generating fully correct code that aligns with user intent. 
    Our study reveals that LLMs tend to pay less attention to user prompts as more code tokens are generated. 
    We hypothesize that this attention dilution issue is an important reason for code generation errors.
    To mitigate this issue, we propose \text{\underline{\textbf{S}}elective \underline{\textbf{P}}rompt \underline{\textbf{A}}nchoring} ({\tool}) to guide code LLMs to pay more attention to user intent when generating code.
    We evaluate {\tool} using six base LLMs across six benchmarks. Our results demonstrate that {\tool} enhances Pass@1 by up to 12.9\%, consistently outperforming SOTA methods in all settings.
     Our code is available at \href{https://github.com/magic-YuanTian/Selective-Prompt-Anchoring}{https://github.com/magic-YuanTian/Selective-Prompt-Anchoring}.
\end{abstract}

\section{Introduction}

\begin{figure*}[!htb]
\vskip 0.3in
    \centering
    \includegraphics[width=\linewidth]{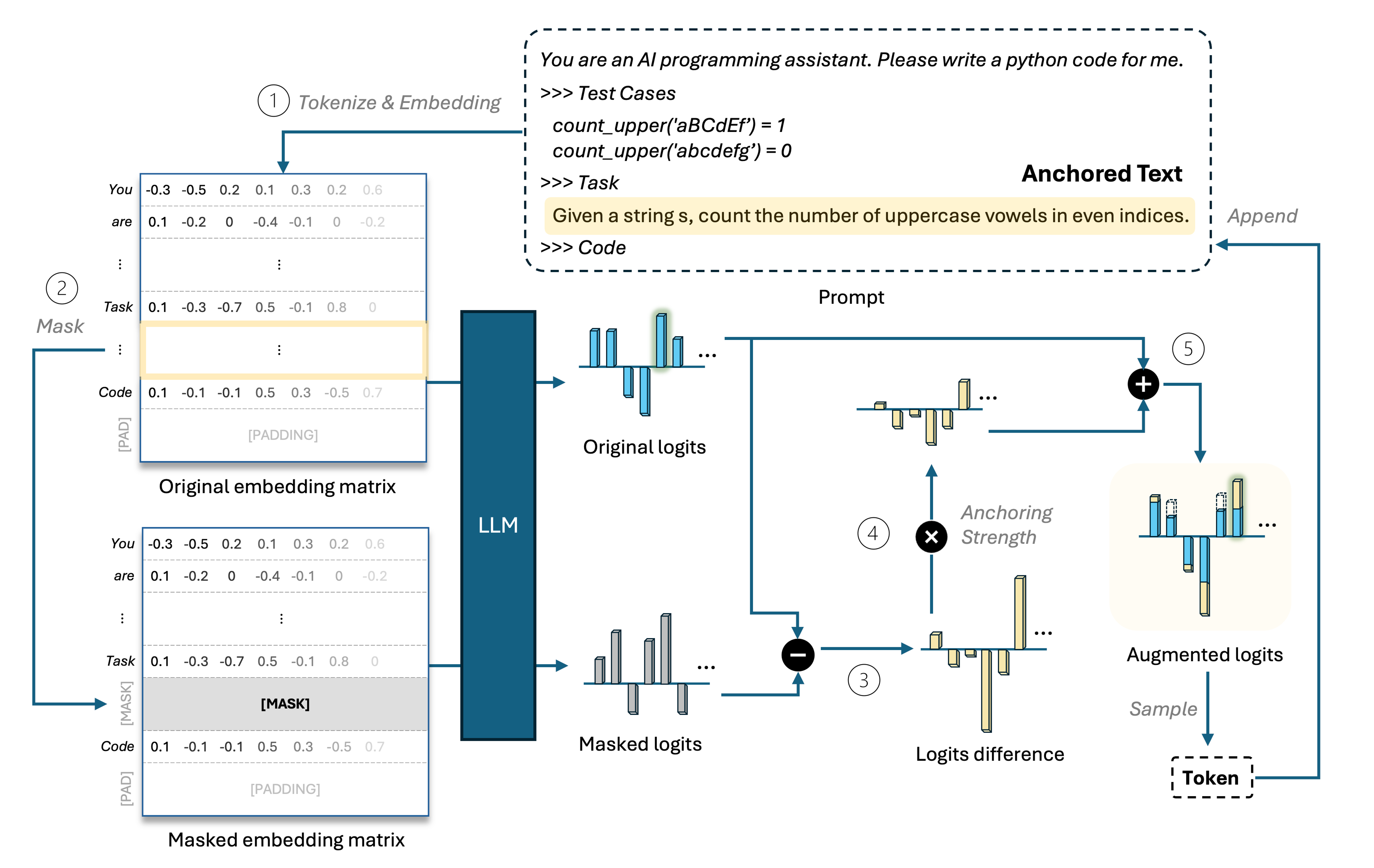}
    \caption{The Workflow of Selective Prompt Anchoring ({\tool}).}
    \label{fig:pipeline}
\end{figure*}

Large language models (LLMs) have emerged as powerful programming assistants. They have demonstrated unprecedented capabilities in interpreting natural language descriptions of programming tasks and generating source code~\cite{deepseek_coder, code_llama}.
Despite this great progress, LLMs still produce incorrect solutions to challenging tasks or generate code that does not fully meet user expectations~\cite{code_gen_error, empricial_iui_sql}. 

To improve the performance of LLMs on coding tasks, many efforts have been made to develop high-quality training data~\citep{starcoder1, deepseek_coder, wei2023magicoder} and design new domain-specific training objectives~\citep{niu2022spt, chakraborty2022natgen}. However, these approaches require tremendous computational resources.
To address this challenge, various prompting methods have been explored to enhance the inference process of code LLMs, e.g., retrieval-augmented generation~\citep{RAG_code1}, chain-of-thoughts~\citep{codechain, cot_code1}, self-planning and debugging~\citep{self_plan, self_debug}, etc. 

Despite these efforts, little is known about why LLMs fail to generate correct code.
In this work, we seek to bridge the knowledge gap by investigating the attention pattern of LLMs during code generation. We analyzed five code LLMs and found that as more code tokens were generated, LLMs paid less attention to the user prompt.
This caused LLMs to gradually deviate from the user intent, thereby leading to code generation errors. 
Furthermore, we found that more generated tokens led to worse code generation performance, demonstrating their struggle with long-term attention.

To mitigate this limitation, we propose \textbf{\underline{S}}elective \textbf{\underline{P}}rompt \textbf{\underline{A}}nchoring ({\tool}), a model-agnostic approach that optimizes LLMs' attention by amplifying the contextual impact of the user prompt.
{\tool} is inspired by the anchoring effect~\citep{anchoring_effect} in psychology, which refers to the phenomenon where humans can be influenced by specific information provided before decision-making.
In {\tool}, we refer to this information as \textit{anchored text}, a group of selected prompt tokens that should receive higher attention from the model than others. Figure~\ref{fig:pipeline} illustrates the workflow of {\tool}.
Given the anchored text, {\tool} creates an original embedding matrix (\circled{1}) as well as a masked embedding matrix by replacing the embeddings corresponding to anchored text with mask embeddings (\circled{2}).
We mathematically show that the anchored text's contextual impact can be approximated by the difference between the logit distribution generated from the original prompt and the prompt with the anchored text masked (\circled{3}).
To amplify the impact of anchored text during code generation, {\tool} multiplies this logit distribution difference by a hyperparameter called \textit{anchoring strength} (\circled{4}), and then adds it to the original logit distribution (\circled{5}).

We evaluate {\tool} on six benchmarks using six code LLMs. The benchmarks cover different programming languages and task difficulty levels, while the LLMs vary in size and code generation performance.
{\tool} enhances Pass@$1$ by up to 12.9\% across all settings, outperforming four SOTA prompting methods and one SOTA attention steering method. Notably, with {\tool}, a smaller version of DeepSeek-Coder (6.7B) can outperform its larger counterpart (33B). 

\section{Attention Analysis of Code LLMs}
\label{sec:empirical}

We conduct an empirical study to investigate the attention dilution phenomenon in code LLMs. 
Following prior studies~\citep{attention1, attention2}, we obtain self-attention scores from the last layer in LLMs, which has been shown to represent the most accurate attention distribution~\citep{model_human_attention_align, last_layer}.\footnote{Intuitively, deeper layers capture representations with long-distance dependencies such as the control flow, which mirrors how humans understand programs. }
We calculate the percentage of attention on the user prompt. Calculation details are provided in Appendix~\ref{app_sub:attention_ratio}. We also experimented with an alternative gradient-based attention calculation method~\citep{gradient1} and obtained similar results, as detailed in Appendix~\ref{app_sub:calculation_gradient_attention}.

\begin{figure}[!h]
    \centering
    \includegraphics[width=\linewidth]{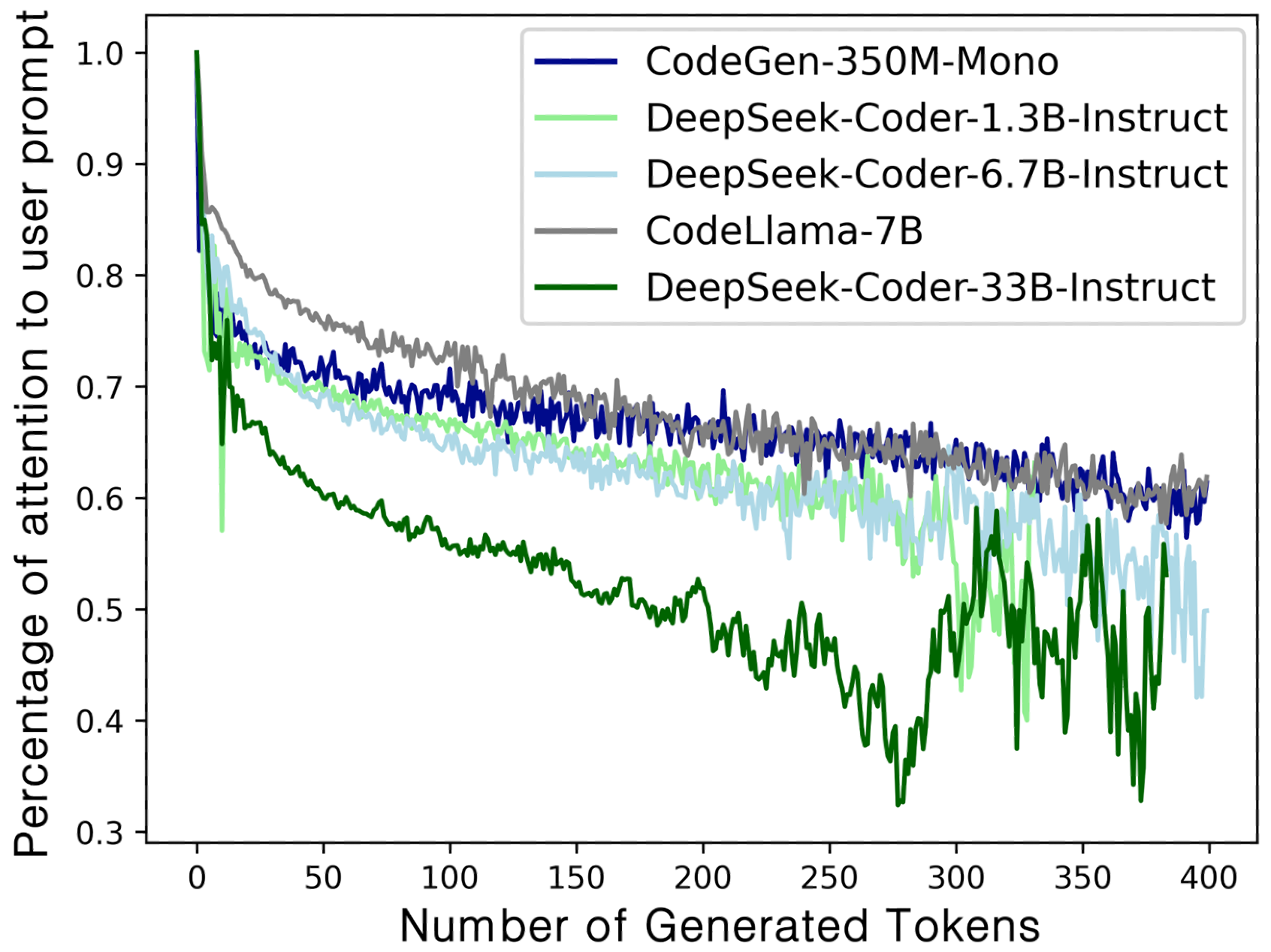}
    \caption{Shift of LLMs' self-attention to the user prompt.}
    \label{fig:self_attention}
\end{figure}

\begin{table}[!h]
\caption{Length of correct vs.~incorrect code generated by LLMs.}
\label{tab:generation_length}
\vskip 0.15in
\begin{center}
\begin{small}
\begin{sc}
\resizebox{0.83\columnwidth}{!}{
\begin{tabular}{l@{\hspace{1.5em}}c@{\hspace{1.5em}}c@{\hspace{1.5em}}c@{\hspace{1.5em}}c}
\toprule
{} & Easy & Medium & Hard & Overall \\
\midrule
Passed & 294 & 475 & 400 & 390 \\
Failed & 418 & 664 & 784 & 622 \\
\bottomrule
\end{tabular}
}
\end{sc}
\end{small}
\end{center}
\vskip -0.1in
\end{table}

We analyzed five code LLMs on HumanEval~\citep{humaneval} and LiveCodeBench~\citep{livecodebench}.
Figure~\ref{fig:self_attention} shows the shift of LLMs' attention on the user prompt during code generation. 
The result shows that as more tokens are generated, models' attention on the user prompt gradually becomes less.
Consequently, the code generation process becomes increasingly influenced by tokens generated in recent time steps, rather than the user prompt. 
This can be problematic in two ways. 
First, any generation inaccuracy, such as creating an unhelpful variable in earlier steps, is likely to propagate and influence subsequent steps.
Second, when generating long and complex code with many detailed requirements, the model may overlook some specifications in the early stages, as the model tends to focus more on recent tokens.
To further investigate whether this phenomenon affects performance, we experimented on LiveCodeBench~\cite{livecodebench}, which provides 3 difficulty levels for each task.
Table~\ref{tab:generation_length} shows that, for tasks with the same difficulty level, the average length of incorrectly generated code is consistently longer than that of correctly generated code. 
This implies that more generation steps can indeed cause attention dilution and hinder accuracy.
More details are included in the Appendix~\ref{app:generation_length_analysis}.

\section{Approach}
\label{sec:approach}

\subsection{Overview}

Given a user prompt $x$, a code LLM ${\model}$ generates tokens autoregressively. At step $i$, the input to ${\model}$ is an $n \times m$ embedding matrix $\mathbf{E}_i$, defined as:
\begin{equation}
\mathbf{E}_i = [\mathbf{E}^x, e_1, e_2, \ldots, e_{i-1}, \texttt{PAD}].
\end{equation}
where $\mathbf{E}^x$ is the submatrix of embeddings for tokens in user prompt $x$, the series $e_1, \ldots, e_{i-1}$ are embeddings of generated tokens, and \texttt{PAD} is a padding submatrix.
The model outputs logits and transforms them into a probability distribution. Then, a sampling method is applied to select the next token.

We propose Selective Prompt Anchoring ({\tool}) to steer model attention by amplifying the contextual impact of the important tokens in the user prompt. In this work, we assume these important tokens are selected by users. 
Inspired by the anchoring effect~\cite{anchoring_effect} in psychology, we call these important tokens ``\textit{anchored text}''.
For example, in the prompt, ``\textit{Given a string, count the number of uppercase vowels}'', the user may want to emphasize ``\textit{uppercase vowels}" to ensure that the model does not forget to define uppercase vowels (i.e., ``\textit{A, E, I, O, U}'') in the generated code. 

In Section~\ref{sec:scale}, we first mathematically model how to steer model attention. Then we augment model attention by increasing the impact of user intent and derive the augmented model output.
In Section~\ref{sec:augmented_logits_approximation}, we derive and calculate the augmented logits, which can be approximately represented by the linear combination of original and masked logits.

\subsection{Attention Steering and Prompt Anchoring}
\label{sec:scale}

{\tool} performs attention steering by scaling the impact of selected tokens to the output logits ${\model}(\mathbf{E}_i)$.
Suppose $x$ is the \textit{anchored text} for which we want to adjust the impact.
\(\mathbf{E}_i\) is an \(n \times m\) input embedding matrix at step $i$, and \(\mathbf{E}^x\) represents a \(n \times k\) submatrix within \(\mathbf{E}_i\) corresponding to the $x$. They are visualized below:
\begin{equation}
\mathbf{E}_i = \left[ \begin{array}{cc}
    \underbrace{\begin{array}{ccc}
    e_{11} & \cdots & e_{1k} \\
    e_{21} & \cdots & e_{2k} \\
    \vdots & \ddots & \vdots \\
    e_{n1} & \cdots & e_{nk}
    \end{array}}_{\mathbf{E}^x}
    &
    \begin{array}{ccc}
    e_{1,k+1} & \cdots & e_{1m} \\
    e_{2,k+1} & \cdots & e_{2m} \\
    \vdots    & \ddots & \vdots \\
    e_{n,k+1} & \cdots & e_{nm}
    \end{array}
\end{array} \right].
\end{equation}
We construct two \(n \times m\) matrices, \(\mathbf{X}\) for user prompt and \(\mathbf{G}_i\) for generated code.
\begin{itemize}
    \item To construct \(\mathbf{X}\), we only retain $k$ columns of \(\mathbf{E}_i\) that correspond to \(\mathbf{E}^x\), while setting remaining columns to zero. This matrix \(\mathbf{X}\) remains constant during token generation.
    \item Conversely, \(\mathbf{G}_i\) is formed by zeroing out the same $k$ columns of \(\mathbf{E}_i\) that correspond to \(\mathbf{E}^x\), while preserving all elements in the remaining columns.
\end{itemize}
They are visualized as follows:
\begin{equation}
    \mathbf{X} = \begin{bmatrix}
    e_{11} & e_{12} & \cdots & e_{1k} & 0 & \cdots & 0 \\
    e_{21} & e_{22} & \cdots & e_{2k} & 0 & \cdots & 0 \\
    \vdots & \vdots & \ddots & \vdots & \vdots & \ddots & \vdots \\
    e_{n1} & e_{n2} & \cdots & e_{nk} & 0 & \cdots & 0 \\
    \end{bmatrix},
\end{equation}
\begin{equation}
    \mathbf{G}_i = \begin{bmatrix}
    0 & 0 & \cdots & 0 & e_{1,k+1} & \cdots & e_{1m} \\
    0 & 0 & \cdots & 0 & e_{2,k+1} & \cdots & e_{2m} \\
    \vdots & \vdots & \ddots & \vdots & \vdots & \ddots & \vdots \\
    0 & 0 & \cdots & 0 & e_{n,k+1} & \cdots & e_{nm} \\
    \end{bmatrix}.
\end{equation}
\(\mathbf{X}\) encapsulates the semantics for the anchored text $x$, while \(\mathbf{G}_i\) encapsulates the semantics for the remaining text, such as the generated code. 
The sum of \(\mathbf{X}\) and \(\mathbf{G}_i\) reconstructs the original matrix \(\mathbf{E}_i\):
\begin{equation}
\mathbf{E}_i = \mathbf{X} + \mathbf{G}_i.
\end{equation}
Suppose we want to scale the semantic impact of the matrix $\mathbf{X}$ by a value ${\weight}$.\footnote{Scaling the semantics of $\mathbf{X}$ by ${\weight}$ is not equivalent to multiplying $\mathbf{X}$ by ${\weight}$. Multiplying the embedding of $\mathbf{X}$ by ${\weight}$ does not simply improve the “semantic influence” of $\mathbf{X}$, since the embedding of $\mathbf{X}$ also encodes other non-semantic information such as positional information. This is why we compute the difference between the logits when masking and unmasking $\mathbf{X}$ to cancel out noise and some of the non-semantic information (detailed in Section~\ref{sec:augmented_logits_approximation}).} We refer to ${\weight}$ as \textit{anchoring strength}.
We use ${\embedding}(\mathbf{X}, {\weight})$ to represent the function that augments ${\embedding}$ by scaling the influence of $\mathbf{X}$ by a degree of ${\weight}$ in the final logit.

\begin{itemize}[noitemsep]
    \item ${\weight} > 1$ indicates semantic amplification, meaning the model generates code with greater consideration of $\mathbf{X}$.
    \item ${\weight} = 1$ indicates using the original embedding. $\mathbf{E}_i$ is equivalent to ${\embedding}(\mathbf{X}, {1})$.
    \item $0 \leq {\weight} < 1$ indicates semantic diminishment, meaning the model generates code with less consideration of $\mathbf{X}$. When ${\weight}$ is 0, the model does not consider $\mathbf{X}$ at all.
    \item ${\weight} < 0$ indicates a reversed semantic impact, meaning the model generates the code in the opposite manner. For example, if $\mathbf{X}$ corresponds to \textit{``uppercase''}, the model will instead consider \textit{``lowercase''}.
\end{itemize}
In this work, we focus on scaling up the impact of the anchored text $x$ to mitigate the attention dilution issue.
Let ${\logits}({\weight})$ represent the augmented logits calculated by model ${\model}$ at step $i$, where the impact of anchored text $x$ is scaled by ${\weight}$.
We can calculate the integral of the partial derivative of ${\model}$ with respect to ${\weight}$ from 0 to ${\weight}$.\footnote{${\model}$ is differentiable for backpropagation.}
Formally, 
\begin{align}
{\logits}({\weight}) &= {\model} ({\embedding}(\mathbf{X}, {\weight}) + \mathbf{G}_i) \\ 
&= {\logits}(0) + \int_0^{\weight} \frac{d{\logits}(t)}{dt} \, dt, \label{eq:10}
\end{align}
where $t$ is the variable of integration.

To reduce computational overhead, attention augmentation is activated when the original generation fails the test case, which is typically included in the code generation context.
When no test case is available, attention augmentation is applied continuously throughout the generation.
In the following section, we explain how to calculate the augmented logits through approximation.

\subsection{Augmented Logits by Approximation}
\label{sec:augmented_logits_approximation}

Given the computational complexities of LLMs, directly solving $\int_0^{\weight} \frac{d{\logits}(t)}{dt} \, dt$ in Equation~\ref{eq:10} is impractical. 
Therefore, we approximate it by employing the Taylor expansion:
\begin{equation}
{\logits}({\weight}) = {\logits}(0) + {\weight} \cdot {\logits}'(0) + \frac{{\weight}^2}{2!}{\logits}''(0) + \ldots     \label{eq:taylor}
\end{equation}
Since LLMs are inherently non-linear, higher-order derivatives of the logits function are non-zero. 
We truncate the Taylor expansion in Equation \ref{eq:taylor} after the first derivative to obtain an approximation, yielding:
\begin{equation}
{\logits}({\weight}) \approx {\logits}(0) + {\weight} \cdot {\logits}'(0),    \label{eq:12}
\end{equation}
where the the integral part $\int_0^{\weight} \frac{d{\logits}(t)}{dt} \, dt$ in Equation~\ref{eq:10} is approximated by ${\weight} \cdot {\logits}'(0)$.
To calculate ${\logits}(0)$,
\footnote{$F_{\theta, i, x}(0)$ does not mean setting the embedding vector to zeros. Instead, it means setting ${\weight}$ to zero, which replaces the original embedding for anchored text with the mask embedding that contains no semantic information. This mask embedding vector is non-zero. Conversely, a zero embedding does not necessarily indicate the absence of semantic information.}
we mask the anchored text $x$ using mask embeddings.
Each LLM provides special tokens reserved for text masking, which almost has no semantic impact,
e.g., \texttt{<unk>} for Code Llama~\citep{code_llama} and \texttt{<pad>} for DeepSeek-Coder~\citep{deepseek_coder}.
Each special token corresponds to a mask embedding. 
By replacing embeddings of $x$ with mask embeddings, we get a masked input matrix $\mathbf{E}^{mask}_i$. It ablates the semantic impact of the anchored text $x$ while ensuring that the positional encoding remains unaffected. 
Thus, we can get 
\begin{equation}
{\logits}(0) = {\model}(\textbf{E}^{mask}_i).    \label{eq:14}
\end{equation}
To calculate ${\logits}'(0)$, we use finite-difference methods to get an approximation. Assuming the interval of $1-0$ is sufficiently small for ${\logits}$, we get:
\begin{equation}
{\logits}'(0) \approx \frac{{\logits}(1)-{\logits}(0)}{1-0}.  \label{eq:15}
\end{equation}
Combining Equations~\ref{eq:12}, \ref{eq:14} and \ref{eq:15}, we get the augmented logits by first-order approximation:
\begin{align}
F_{\theta,i,x}(\omega) &\approx F_{\theta,i,x}(0) + \omega \cdot (F_{\theta,i,x}(1) - F_{\theta,i,x}(0)) \label{eq:14} \\
&= \omega \cdot f_\theta({\embedding}(\mathbf{X},1) + \mathbf{G}_i) \notag \\
&\quad + (1-\omega) \cdot f_\theta({\embedding}(\mathbf{X},0) + \mathbf{G}_i) \label{eq:15} \\
&= \omega \cdot f_\theta(\mathbf{E}_i) + (1-\omega) \cdot f_\theta(\mathbf{E}_i^{mask}). \label{eq:17}
\end{align}
Based on the augmented logits ${\logits}({\weight})$ where the impact of the anchored text is scaled by the value ${\weight}$, a certain sampling algorithm (e.g., greedy sampling) can be applied to select the token.
We provide more discussion about approximation in Appendix~\ref{app:Higher_order_approximation}.

We choose not to directly modify self-attention layers to steer the model attention, since this requires identifying which attention head in which layer to steer and the direct editing of attention values does not synchronize with other components like the feedforward layers, which can be fairly brittle and costly. Instead, we chose to simulate attention steering by logits manipulation, which is fast, reliable, and model-agnostic. In the next section, we demonstrate that {\tool} achieves better performance with less computational overhead compared to a SOTA method that directly modifies self-attention layers~\cite{pasta}. 

\section{Experiment Setup}
\label{sec:exp}

\subsection{Benchmarks}
\textbf{HumanEval}~\citep{humaneval} includes 164 Python tasks designed by OpenAI and now has become a widely-used benchmark for code generation.

\textbf{MBPP}~\citep{mbpp} is another popular benchmark that includes 974 crowd-sourced Python tasks.
Due to ambiguous task descriptions, the authors of MBPP created a sanitized version that included 427 tasks with clearer descriptions. 
We evaluate {\tool} on the sanitized version.

\textbf{HumanEval+} and \textbf{MBPP+}~\citep{evalplus} improves the original HumanEval and MBPP benchmarks with additional test cases to cover corner cases~\citep{code_reliability}.

\textbf{HumanEval-X} \cite{humaneval_x} extends the HumanEval benchmark to support more programming languages such as Python, Java, JavaScript, C++, and Go. It aims to evaluate the multilingual code generation abilities.

\textbf{BigCodeBench} \cite{bigcodebench} is a more challenging benchmark for code generation that evaluates models' abilities to follow complex instructions and use tools, including 1,140 real-world Python tasks across 139 libraries.

\textbf{LiveCodeBench} \cite{livecodebench} is a contamination-free benchmark sourced from competitive programming platforms.  
The benchmark continues to evolve and add new code generation tasks. We conducted experiments on the latest release, which includes a total of 1,055 tasks. The latest tasks span from October 1, 2024, to February 1, 2025.

\subsection{Baselines}
\textbf{Base Models.}
We select six representative open-source code LLMs: CodeGen-Mono-350M~\cite{codegen}, CodeLlama-7B~\cite{code_llama}, StarCoder2-15B~\cite{starcoder2}, and DeepSeek-Coder-Instruct-1.3B, 6.7B, and 33B~\cite{deepseek_coder}. 
These models exhibit varying levels of performance in code generation.
We include more setup details about models in Appendix~\ref{app:model_set_up}.

\textbf{Attention Steering Baseline.} PASTA~\cite{pasta} is a recent method designed to steer model attention for better instruction following. 
Unlike {\tool}, PASTA requires a model-specific and time-consuming profiling stage to identify attention headers that are beneficial to performance. 
For each unique task, PASTA requires around 1000 training samples to identify attention headers that are beneficial to performance.
By contrast, {\tool} is model-agnostic and only needs tuning a single hyperparameter with a few samples, which is fast and generalizable as detailed in Appendix~\ref{app:tuning_study}.
PASTA internally edits transformers' self-attention during the feed-forward process, whereas {\tool} only edits the final logits based on a mathematical approximation.

\textbf{Prompting Methods.} In addition to PASTA, we also compare {\tool} to the mainstream prompting-based code generation methods, including Self-Debugging~\cite{self_debug}, Self-Planning~\cite{self_plan}, ReAct~\cite{react}, and Self-Edit~\cite{self_edit}. Self-Debugging and Self-Edit leverage error messages from test cases to refine generated code. Self-planning generates a step-by-step plan before code generation. ReAct prompts an LLM to generate reasoning traces and action plans in an interleaved manner.

\subsection{Evaluation Metrics} 
Following prior work \cite{humaneval}, we measure code generation performance using the Pass@$k$ metric, which measures whether any of the top $k$ candidates can pass all the test cases.
For main results, we report Pass@$1$ using greedy sampling to generate a single code snippet.
To demonstrate the generalizability of {\tool}, we further calculate Pass@$10$ using beam search in Appendix~\ref{app:beam_search}.

\subsection{Anchoring Strength Tuning}
\label{sec:weight_approach}
The anchoring strength ${\weight}$ serves as a hyperparameter in {\tool}.
For each model and dataset, we use grid search to tune the anchoring strength ${\weight}$ on 1/5 of the tasks, and evaluate Pass@1 of {\tool} on the remaining 4/5 of the tasks. This process is repeated across all five folds, with final performance metrics averaged across folds.\footnote{We tune {\tool} on all tasks for better generalizability, while {\tool} is only activated for failed tasks during inference to optimize computational efficiency and generation performance.}
We observe an unimodal relationship between ${\weight}$ and the performance.
We provide more details for the tuning {\tool} in Appendix~\ref{app:tuning_study}.

\section{Results}
\label{sec:results}

\begin{table*}[h!]
    \centering
    \caption{Absolute ($\Delta$) and Relative ($\uparrow$) Performance improvements in Pass@$1$ rates (\%).}
    \vspace{0.5\baselineskip}
    \label{tab:results}
    \begin{center}
     \begin{small}
     \begin{sc}
     \resizebox{\linewidth}{!}{
    \begin{tabular}{
        l
        c
        l
        l
        l
        l
        l
        l 
        }
        \toprule
        Model & {\footnotesize Size} & {HumanEval} & {HumanEval+} & {MBPP} & {MBPP+} & {BigCodeBench} & {LiveCodeBench} \\
        \midrule
        CodeGen-Mono & {\moremoresmall (350M)} & 15.3 & 12.2 & 19.6 & 15.9 & 1.1 & 1.1 \\
        + {\tool} & & 20.1\!\raisebox{2pt}{\textcolor{mygreen}{\specialtiny$\begin{array}{l}\Delta\!+\!4.8\\(31\%\uparrow)\end{array}$}} & 17.1\!\raisebox{2pt}{\textcolor{mygreen}{\specialtiny$\begin{array}{l}\Delta\!+\!4.9\\(40\%\uparrow)\end{array}$}} & 27.4\!\raisebox{2pt}{\textcolor{mygreen}{\specialtiny$\begin{array}{l}\Delta\!+\!7.8\\(40\%\uparrow)\end{array}$}} & 22.6\!\raisebox{2pt}{\textcolor{mygreen}{\specialtiny$\begin{array}{l}\Delta\!+\!6.7\\(42\%\uparrow)\end{array}$}} & 1.6\!\raisebox{2pt}{\textcolor{mygreen}{\specialtiny$\begin{array}{l}\Delta\!+\!0.5\\(45\%\uparrow)\end{array}$}} & 1.5\!\raisebox{2pt}{\textcolor{mygreen}{\specialtiny$\begin{array}{l}\Delta\!+\!0.4\\(36\%\uparrow)\end{array}$}} \\
        \specialrule{.02pt}{2pt}{2pt}
        DeepSeek-Coder & {\moremoresmall (1.3B)} & 66.4 & 61.8 & 58.2 & 52.4 & 2.5 & 6.5 \\
        + {\tool} & & 70.1\!\raisebox{2pt}{\textcolor{mygreen}{\specialtiny$\begin{array}{l}\Delta\!+\!3.7\\(6\%\uparrow)\end{array}$}} & 67.7\!\raisebox{2pt}{\textcolor{mygreen}{\specialtiny$\begin{array}{l}\Delta\!+\!5.9\\(10\%\uparrow)\end{array}$}} & 60.9\!\raisebox{2pt}{\textcolor{mygreen}{\specialtiny$\begin{array}{l}\Delta\!+\!2.7\\(5\%\uparrow)\end{array}$}} & 52.4\!\raisebox{2pt}{\textcolor{lightgray}{\specialtiny$\begin{array}{l}\Delta\!+\!0.0\\(0\%\uparrow)\end{array}$}} & 3.4\!\raisebox{2pt}{\textcolor{mygreen}{\specialtiny$\begin{array}{l}\Delta\!+\!0.9\\(36\%\uparrow)\end{array}$}} & 9.2\!\raisebox{2pt}{\textcolor{mygreen}{\specialtiny$\begin{array}{l}\Delta\!+\!2.7\\(42\%\uparrow)\end{array}$}} \\
        \specialrule{.02pt}{2pt}{2pt}
        {DeepSeek-Coder} & {\moremoresmall (6.7B)} & 75.6 & 70.2 & 67.0 & 58.5 & 12.7 & 7.8 \\
        + {\tool} & & 88.5\!\raisebox{2pt}{\textcolor{mygreen}{\specialtiny$\begin{array}{l}\Delta\!+\!12.9\\(17\%\uparrow)\end{array}$}} & 79.9\!\raisebox{2pt}{\textcolor{mygreen}{\specialtiny$\begin{array}{l}\Delta\!+\!9.7\\(14\%\uparrow)\end{array}$}} & 71.0\!\raisebox{2pt}{\textcolor{mygreen}{\specialtiny$\begin{array}{l}\Delta\!+\!4.0\\(6\%\uparrow)\end{array}$}} & 60.7\!\raisebox{2pt}{\textcolor{mygreen}{\specialtiny$\begin{array}{l}\Delta\!+\!2.2\\(4\%\uparrow)\end{array}$}} & 16.4\!\raisebox{2pt}{\textcolor{mygreen}{\specialtiny$\begin{array}{l}\Delta\!+\!3.7\\(29\%\uparrow)\end{array}$}} & 10.8\!\raisebox{2pt}{\textcolor{mygreen}{\specialtiny$\begin{array}{l}\Delta\!+\!3.0\\(38\%\uparrow)\end{array}$}} \\
        \specialrule{.02pt}{2pt}{2pt}
        {CodeLlama} & {\moremoresmall (7B)} & 33.6 & 28.2 & 50.9 & 40.8 & 3.4 & 3.8 \\
        + {\tool} & & 44.0\!\raisebox{2pt}{\textcolor{mygreen}{\specialtiny$\begin{array}{l}\Delta\!+\!10.4\\(31\%\uparrow)\end{array}$}} & 36.0\!\raisebox{2pt}{\textcolor{mygreen}{\specialtiny$\begin{array}{l}\Delta\!+\!7.8\\(28\%\uparrow)\end{array}$}} & 54.3\!\raisebox{2pt}{\textcolor{mygreen}{\specialtiny$\begin{array}{l}\Delta\!+\!3.4\\(7\%\uparrow)\end{array}$}} & 44.0\!\raisebox{2pt}{\textcolor{mygreen}{\specialtiny$\begin{array}{l}\Delta\!+\!3.2\\(8\%\uparrow)\end{array}$}} & 4.1\!\raisebox{2pt}{\textcolor{mygreen}{\specialtiny$\begin{array}{l}\Delta\!+\!0.7\\(5\%\uparrow)\end{array}$}} & 4.0\!\raisebox{2pt}{\textcolor{mygreen}{\specialtiny$\begin{array}{l}\Delta\!+\!0.2\\(5\%\uparrow)\end{array}$}} \\
        \specialrule{.02pt}{2pt}{2pt}
        {StarCoder2} & {\moremoresmall (16B)} & 67.7 & 60.4 & 78.0 & 65.1 & 13.3 & 7.0 \\
        + {\tool} & & 75.6\!\raisebox{2pt}{\textcolor{mygreen}{\specialtiny$\begin{array}{l}\Delta\!+\!7.9\\(12\%\uparrow)\end{array}$}} & 65.6\!\raisebox{2pt}{\textcolor{mygreen}{\specialtiny$\begin{array}{l}\Delta\!+\!5.2\\(9\%\uparrow)\end{array}$}} & 82.0\!\raisebox{2pt}{\textcolor{mygreen}{\specialtiny$\begin{array}{l}\Delta\!+\!4.0\\(5\%\uparrow)\end{array}$}} & 69.1\!\raisebox{2pt}{\textcolor{mygreen}{\specialtiny$\begin{array}{l}\Delta\!+\!4.0\\(6\%\uparrow)\end{array}$}} & 14.3\!\raisebox{2pt}{\textcolor{mygreen}{\specialtiny$\begin{array}{l}\Delta\!+\!1.0\\(8\%\uparrow)\end{array}$}} & 8.2\!\raisebox{2pt}{\textcolor{mygreen}{\specialtiny$\begin{array}{l}\Delta\!+\!1.2\\(17\%\uparrow)\end{array}$}} \\
        \specialrule{.02pt}{2pt}{2pt}
        {DeepSeek-Coder} & {\moremoresmall (33B)} & 81.7 & 77.1 & 73.4 & 63.2 & 18.9 & 11.9 \\
        + {\tool} & & 86.2\!\raisebox{2pt}{\textcolor{mygreen}{\specialtiny$\begin{array}{l}\Delta\!+\!4.5\\(6\%\uparrow)\end{array}$}} & 79.3\!\raisebox{2pt}{\textcolor{mygreen}{\specialtiny$\begin{array}{l}\Delta\!+\!2.2\\(3\%\uparrow)\end{array}$}} & 79.4\!\raisebox{2pt}{\textcolor{mygreen}{\specialtiny$\begin{array}{l}\Delta\!+\!6.0\\(8\%\uparrow)\end{array}$}} & 70.3\!\raisebox{2pt}{\textcolor{mygreen}{\specialtiny$\begin{array}{l}\Delta\!+\!7.1\\(11\%\uparrow)\end{array}$}} & 21.5\!\raisebox{2pt}{\textcolor{mygreen}{\specialtiny$\begin{array}{l}\Delta\!+\!2.6\\(14\%\uparrow)\end{array}$}} & 15.8\!\raisebox{2pt}{\textcolor{mygreen}{\specialtiny$\begin{array}{l}\Delta\!+\!3.9\\(33\%\uparrow)\end{array}$}} \\
        \bottomrule
    \end{tabular}
    }
   \end{sc}
  \end{small}
 \end{center}
 \vskip -0.1in
\end{table*}

\subsection{Improvement over Base Models}
\label{sec:main_results}

Table~\ref{tab:results} shows {\tool} consistently improves Pass@1 across all 6 benchmarks and 6 code LLMs.
On HumanEval/HumanEval+ and MBPP/MBPP+, {\tool} enhances the base model with on average 5.5\% absolute improvement and 14.5\% relative improvement, achieving up to a 12.9\% absolute improvement and a 42\% relative improvement.
These improvements are observed across models of varying sizes (350M-33B), original performance (15\%-86\%), and architectures.
Notably, with {\tool}, the smaller DeepSeek-Coder (6.7B) outperforms its much larger 33B counterpart on HumanEval. 
This suggests optimizing model attention is a promising alternative to simply scaling up model size~\cite{scaling_law} for performance improvement.
On BigCodeBench and LiveCodeBench, while the absolute improvements (average 1.57\% and 1.90\%) are more modest compared to other benchmarks, the relative gains (average 22.83\% and 28.50\%) remain significant.
This is because {\tool} leverages the ability of its base model. If the base model could solve a task but overlooks a few important tokens in the prompt, {\tool} can help with this by adjusting the attention. If a model lacks the ability to solve a task, adjusting the model attention will not help much.

To demonstrate the generalizability of {\tool}, we show that {\tool} can consistently improve Pass@10 by, on average 3.42\% and up to 7.9\%, as detailed in Appendix~\ref{app:beam_search}.
We further evaluate {\tool} in scenarios where test cases are not available and show that it still significantly enhances performance, achieving an average 4.7\% Pass@1 improvement on HumanEval, as detailed in Appendix~\ref{app:no_test_case}.
We illustrate {\tool}'s attention anchoring with two concrete examples in Appendix~\ref{app:examples}.
{\tool} improves performance by simply aligning attention to prompts, without introducing additional model parameters or context.
We believe this success comes from its ability to improve attention reliability.
We provide a thorough discussion in Appendix~\ref{app:explanation}.

\subsection{Comparison to SOTA Methods}

\textbf{PASTA.} 
Table~\ref{tab:prompt_optimization_results} shows that {\tool} outperforms PASTA by achieving a 5.9X higher Pass@1 improvement while using only 20\% of the inference time.\footnote{For a fair comparison, we average and add the latency of PASTA's model profiling and the {\tool}'s tuning to the inference time for each task.}
We include more experimental details and discussion in Appendix~\ref{app:comparison_to_pasta}.
Compared to the base model, {\tool} increases the decoding time by a practically negligible factor of 1.27. Further discussion about the computational cost of {\tool} can be found in Appendix~\ref{app:computation}.

\textbf{Prompting Methods.} 
Table~\ref{tab:prompt_optimization_results} shows that {\tool} outperforms Self-Debugging, Self-Edit, Self-Planning, and ReAct by achieving improvements that are 1.8X, 4.3X, 2.1X, and 5.9X higher, respectively, while only using 36\%, 37\%, 45\%, and 34\% of the inference time.
The performance superiority stems from two key factors. 
On the one hand, unlike prompting methods that add more information or enforce generation workflows, {\tool} preserves the original prompt and does not involve additional LLM calls, thereby achieving better time efficiency. 
On the other hand, {\tool} addresses the attention dilution issue that prompting methods do not explicitly handle, thereby achieving higher accuracy.

\begin{table}[!h]
\caption{Comparison between {\tool} and SOTA methods.}
\label{tab:prompt_optimization_results}
\vskip 0.15in
\begin{center}
\begin{small}
\begin{sc}
\resizebox{0.87\columnwidth}{!}{
\begin{tabular}{l@{\hspace{1.5em}}c@{\hspace{1.5em}}c}
\toprule
{Method} & {$\Delta$Pass@1 (\%)} & {Time (Sec)} \\
\midrule
Base Model & 0 & 7.7 \\
\midrule
Pasta & +1.2 & 48.8 \\
Self-Debugging & +4.2 & 27.3 \\
Self-Edit & +1.8 & 26.4 \\
Self-Planning & +3.6 & 21.6 \\
ReAct & +1.3 & 28.8 \\
{\tool} & \textbf{+7.7} & \textbf{9.8} \\
\bottomrule
\end{tabular}
}
\end{sc}
\end{small}
\end{center}
\vskip -0.1in
\end{table}

\subsection{Evaluation on Different Programming Languages}

To evaluate the generalizability across different programming languages, we further evaluate {\tool} using HumanEval-X~\cite{humaneval_x}, which includes five programming languages.\footnote{In the latest version, Rust caused issues when running test cases, so we excluded Rust from the results.}
Table~\ref{tab:humaneval_x_results} demonstrates that {\tool} consistently improves Pass@1 on HumanEval-X, with an average increase of 7.9\% for Python, 4.85\% for Java, 6.5\% for JavaScript, 3.65\% for C++, and 5.2\% for Go.

\begin{table}[h]
    \caption{Evaluation on HumanEval-X with Different Languages.}
    \label{tab:humaneval_x_results}
    \vskip 0.15in
    \begin{center}
    \begin{small}
    \begin{sc}
    \resizebox{0.87\columnwidth}{!}{
    \begin{tabular}{l@{\hspace{3pt}}c@{\hspace{6pt}}c@{\hspace{6pt}}c@{\hspace{6pt}}c@{\hspace{6pt}}c}
    \toprule
    Model & Py & Java & JS & C++ & Go \\
    \midrule
    Codegen \smallerfootnote{(350M)} & 15.3 & 9.8 & 13.4 & 9.8 & 6.7 \\
    +{\tool} & {\moremoresmall \textcolor{mygreen}{+4.8}} & {\moremoresmall \textcolor{mygreen}{+3.0}} & {\moremoresmall \textcolor{mygreen}{+4.3}} & {\moremoresmall \textcolor{mygreen}{+3.7}} & {\moremoresmall \textcolor{mygreen}{+6.7}} \\
    \midrule
    DeepSeek \smallerfootnote{(1.3B)} & 66.4 & 42.7 & 57.3 & 43.3 & 40.2 \\
    +{\tool} & {\moremoresmall \textcolor{mygreen}{+6.7}} & {\moremoresmall \textcolor{mygreen}{+3.0}} & {\moremoresmall \textcolor{mygreen}{+3.0}} & {\moremoresmall \textcolor{mygreen}{+2.4}} & {\moremoresmall \textcolor{mygreen}{+1.8}} \\
    \midrule
    DeepSeek \smallerfootnote{(6.7B)} & 75.6 & 48.8 & 65.2 & 49.4 & 45.7 \\
    +{\tool} & {\moremoresmall \textcolor{mygreen}{+12.9}} & {\moremoresmall \textcolor{mygreen}{+8.5}} & {\moremoresmall \textcolor{mygreen}{+11.6}} & {\moremoresmall \textcolor{mygreen}{+1.8}} & {\moremoresmall \textcolor{mygreen}{+7.3}} \\
    \midrule
    CodeLlama \smallerfootnote{(7B)} & 33.6 & 22.0 & 29.3 & 22.0 & 20.1 \\
    +{\tool} & {\moremoresmall \textcolor{mygreen}{+10.4}} & {\moremoresmall \textcolor{mygreen}{+6.1}} & {\moremoresmall \textcolor{mygreen}{+8.5}} & {\moremoresmall \textcolor{mygreen}{+6.1}} & {\moremoresmall \textcolor{mygreen}{+6.7}} \\
    \midrule
    StarCoder2 \smallerfootnote{(15B)} & 67.7 & 22.0 & 29.3 & 22.0 & 20.1 \\
    +{\tool} & {\moremoresmall \textcolor{mygreen}{+7.9}} & {\moremoresmall \textcolor{mygreen}{+5.5}} & {\moremoresmall \textcolor{mygreen}{+7.9}} & {\moremoresmall \textcolor{mygreen}{+5.5}} & {\moremoresmall \textcolor{mygreen}{+6.1}} \\
    \midrule
    DeepSeek \smallerfootnote{(33B)} & 81.7 & 53.0 & 70.7 & 53.7 & 49.4 \\
    +{\tool} & {\moremoresmall \textcolor{mygreen}{+4.5}} & {\moremoresmall \textcolor{mygreen}{+3.0}} & {\moremoresmall \textcolor{mygreen}{+3.7}} & {\moremoresmall \textcolor{mygreen}{+2.4}} & {\moremoresmall \textcolor{mygreen}{+2.4}} \\
    \bottomrule
    \end{tabular}
    }
    \end{sc}
    \end{small}
    \end{center}
    \vskip -0.1in
\end{table}

\subsection{Ablation Study of Anchored Text Selection}
\label{sec:anchor_ablation}

To investigate the impact of anchored text selection in code generation tasks, we conduct an ablation study by masking different components in code generation prompts.
We decompose the code generation prompt into three components: (1) \textit{natural language description}, (2) \textit{source code}, and (3) \textit{test cases}.
We create 4 automated anchored text selection methods by ablating the source code and test cases.\footnote{We do not ablate the natural language description since it explicitly represents the user intent.}
To explore whether anchoring on a few more informative tokens can enhance performance, we further create a condition using important tokens labeled by human programmers as the anchored text.
We leverage the dataset from \citet{model_human_attention_align}, where multiple programmers manually identified critical tokens that models should attend to when solving HumanEval and MBPP programming tasks. Following our previous experimental setup, we tune {\tool} for each condition and benchmark, calculating the average Pass@$1$ improvement across all 6 experimental LLMs.

\begin{table}[h]
    \centering
    \caption{Pass@1 Improvement (\%) with Different Anchored Texts (\textsc{He}: HumanEval, \textsc{Mb}: MBPP, \textsc{Bcb}: BigCodeBench, \textsc{Lcb}: LiveCodeBench).}
    \label{tab:selective_prompt_ablation}
    \vskip 0.15in
    \begin{center}
    \begin{small}
    \begin{sc}
    \resizebox{\columnwidth}{!}{ 
    \begin{tabular}{l@{\hspace{7pt}}*{6}{l@{\hspace{7pt}}}}  
        \toprule
        Anchored Text & He & He+ & Mb & Mb+ & Bcb & Lcb \\ 
        \midrule
        \begin{tabular}[c]{@{}l@{}} Human Attention \\ ~\cite{model_human_attention_align} \end{tabular} & 3.66 & 3.04 & 2.58 & 2.11 & n/a & n/a \\ 
        \midrule
        Natural Language & \textbf{5.48} & \textbf{5.08} & \textbf{4.26} & \textbf{3.22} & \textbf{1.57} & \textbf{1.90} \\
        + Code & 4.87 & 4.65 & 3.75 & 2.81 & 1.42 & 1.73 \\ 
        + Test Cases & 5.11 & 4.89 & 4.05 & 3.11 & 1.50 & 1.82 \\ 
        + Code \& Test Cases & 4.76 & 4.57 & 3.98 & 2.81 & 1.44 & 1.73 \\
        \bottomrule
    \end{tabular}
    }
    \end{sc}
    \end{small}
    \end{center}
    \vskip -0.1in
\end{table}

As shown in Table~\ref{tab:selective_prompt_ablation}, anchoring the natural language description alone achieves the best performance improvement.
For example, on HumanEval, anchoring the natural language description alone improves Pass@1 by 5.48\%. Including source code (4.87\%) and test cases (5.11\%) in the anchored text make the performance worse. When anchoring the entire user prompt, performance deteriorates further to 4.76\%.
This suggests that removing less relevant context can help {\tool} better align LLM attention with the user intent defined in the natural language description.

Interestingly, we find that anchoring using human-labeled important tokens achieves only 3.66\% on HumanEval, which is consistently less effective than all the automated experimental anchoring methods.
It suggests that narrowing the anchored text to more informative tokens does not necessarily improve performance.
We think there are two plausible reasons. First, since LLMs need to attend to different context tokens at each decoding step, providing a narrow set of anchored tokens may have a negative impact and distract the LLM in certain decoding steps. 
Second, previous studies such as~\citet{attention_sink} show that even though some tokens, such as separators and empty space, may not be semantically meaningful or informative, they provide important signals for LLMs to generate the right content (e.g., following the grammar rules). Thus, over-attending to the informative tokens but not the special tokens in the task description may disrupt the regular generation process.
Nevertheless, we think this is a challenging but interesting future direction to investigate. We believe our findings will open up new research opportunities for the community on this topic.

\subsection{Impact of Prompt Length}
\label{sec:prompt_length_analysis}

To further validate that {\tool} effectively addresses the attention dilution issue, we analyze both the original model performance and {\tool}'s effectiveness across different prompt lengths, as shown in Table~\ref{tab:prompt_length_results}.
We divide the HumanEval dataset into three equal-sized subsets (\textit{Short}, \textit{Medium}, and \textit{Long}) based on the 33$^{\rm rd}$ and 66$^{\rm th}$ percentiles of prompt lengths. 
We find that the original code LLMs consistently achieve better performance on shorter prompts compared to longer ones. The average Pass@1 is 74.3\% for \textit{Short} prompts, 49.2\% for \textit{Medium} prompts, and only 30.8\% for \textit{Long} prompts. 
This result echoes our empirical finding on attention dilution, as longer prompts can lead to more severe attention dilution issues.

We further investigate whether {\tool} can effectively address the attention dilution issue by comparing performance improvements across three subsets of tasks, \textit{Short}, \textit{Medium}, and \textit{Long}.
To ensure a fair comparison, the attention augmentation of {\tool} is activated for all tasks in each subset of this experiment.
The average Pass@1 improvement of {\tool} is 1.1\% for \textit{Short}, 3.2\% for \textit{Medium}, and 11.7\% for \textit{Long}.
It shows that {\tool} consistently achieves greater improvements for longer prompts compared to shorter ones.
This implies that {\tool} indeed improves performance by addressing the attention dilution issue, as the improvement is more significant when attention dilution is more severe.

\begin{table}[h]
    \caption{{\tool}'s performance on tasks with different prompt lengths.}
    \label{tab:prompt_length_results}
    \vskip 0.15in
    \begin{center}
    \begin{small}
    \begin{sc}
    \resizebox{0.84\columnwidth}{!}{
    \begin{tabular}{l@{\hspace{0.5em}}c@{\hspace{0.5em}}c@{\hspace{0.5em}}c}
    \toprule
    {Model} & {Short} & {Medium} & {Long} \\
    \midrule
    CodeGen \smallerfootnote{(350M)} & 37.0 & 6.6& 2.3 \\
    + {\tool} & {\moremoresmall \textcolor{mygreen}{+1.7}} & {\moremoresmall \textcolor{mygreen}{+3.0}} & {\moremoresmall \textcolor{mygreen}{+4.3}} \\
    \midrule
    DeepSeek-Coder \smallerfootnote{(1.3B)} & 81.8 & 45.5 & 29.6 \\
    + {\tool} & {\moremoresmall \textcolor{myred}{-1.8}} & {\moremoresmall \textcolor{mygreen}{+14.5}} & {\moremoresmall \textcolor{mygreen}{+26.0}} \\
    \midrule
    DeepSeek-Coder \smallerfootnote{(6.7B)} & 87.3 & 69.1 & 44.4 \\
    + {\tool} & {\moremoresmall \textcolor{mygreen}{+3.6}} & {\moremoresmall \textcolor{myred}{-3.6}} & {\moremoresmall \textcolor{mygreen}{+7.5}} \\
    \midrule
    CodeLlama \smallerfootnote{(7B)} & 69.2 & 43.5 & 0 \\
    + {\tool} & {\moremoresmall \textcolor{mygreen}{+2.6}} & {\moremoresmall \textcolor{mygreen}{+0.0}} & {\moremoresmall \textcolor{mygreen}{+10.0}} \\
    \midrule
    StarCoder2 \smallerfootnote{(15B)} & 85.5 & 49.1 & 40.1 \\
    + {\tool} & {\moremoresmall \textcolor{myred}{-1.8}} & {\moremoresmall \textcolor{mygreen}{+5.5}} & {\moremoresmall \textcolor{mygreen}{+14.8}} \\
    \midrule
    DeepSeek-Coder \smallerfootnote{(33B)} & 85.5 & 81.8 & 68.5 \\
    + {\tool} & {\moremoresmall \textcolor{mygreen}{+1.8}} & {\moremoresmall \textcolor{mygreen}{+0.0}} & {\moremoresmall \textcolor{mygreen}{+7.4}} \\
    \bottomrule
    \end{tabular}
    }
    \end{sc}
    \end{small}
    \end{center}
    \vskip -0.1in
\end{table}

\subsection{Impact of Anchoring Strength}
\label{sec:strength_impact}

\begin{figure*}[h]
    \centering
    \includegraphics[width=1\linewidth]{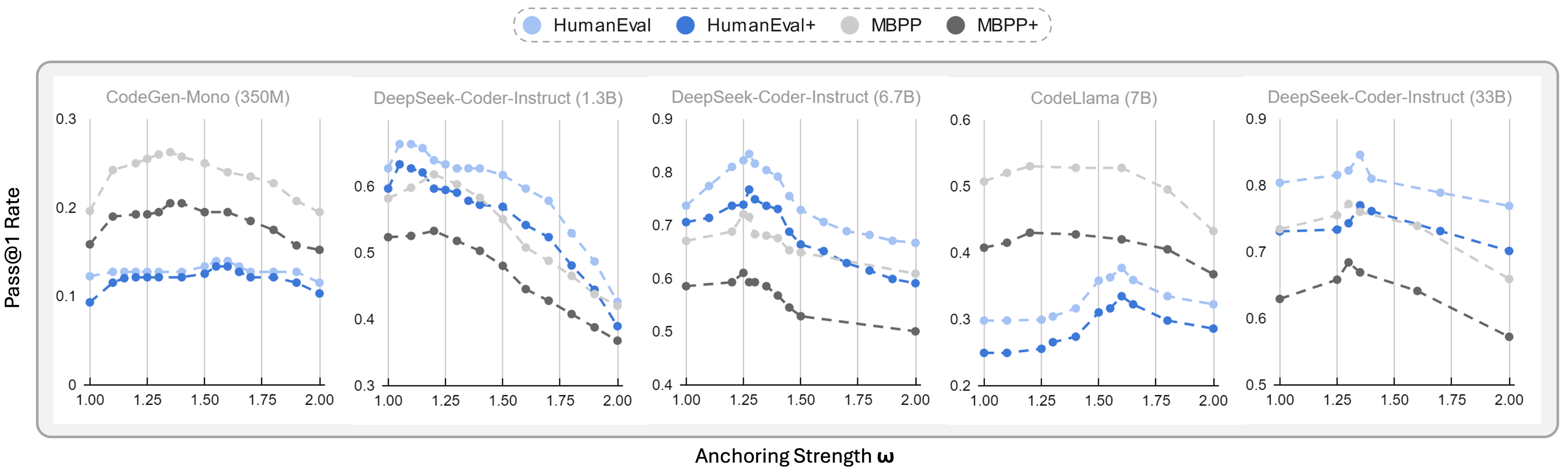}
    \caption{Analysis of Anchoring Strength}
    \label{fig:weight_analysis}
\end{figure*}

In this section, we discuss how the anchoring strength ${\weight}$ of {\tool} influences performance. 
We observe that different values of ${\weight}$ yield different effects: when ${\weight}$ is too low, performance gains are limited; when it is too high, the LLM becomes biased, leading to performance degradation.
As shown in Figure~\ref{fig:weight_analysis}, ${\weight}=1$ corresponds to the original generation. As ${\weight}$ increases, performance initially improves, reaches an optimum, and subsequently declines with further increases in ${\weight}$ . This simple pattern applies to all the settings.
We find ${\weight}$ is both tunable and transferable. While the optimal value may vary slightly across models and benchmarks, it is generally model-dependent. A value of ${\weight}=1.25$ consistently yields robust improvements across all settings. Further details are provided in Appendix~\ref{app:tuning_study}.

To make {\tool} more stable and eliminate the need of tuning, we also propose a confidence-modulated weighting strategy that adjusts each token’s anchoring strength based on its original probability. Details are provided in Appendix~\ref{app:confidence_modulate}.

\subsection{Evaluation on Other Generative Tasks}
\label{sec:other_generative_tasks}

\begin{table}[h]
    \caption{{\tool}'s performance on different generative tasks (``\textsc{Truth.}'' and ``\textsc{Info.}'' are evaluation metrics of TruthfulQA).}
    \vskip 0.15in
    \begin{center}
    \begin{small}
    \begin{sc}
    \resizebox{0.87\columnwidth}{!}{
    \begin{tabular}{@{}l@{\hspace{4pt}}c@{\hspace{4pt}}c@{\hspace{4pt}}c@{\hspace{4pt}}c@{\hspace{4pt}}c@{}}
    \toprule
    Model & Truth. & Info. & GSM8K & MMLU & BoolQ \\
    \midrule
    Llama-3.1 & 88.4 & 97.8 & 76.5& 69.5 & 83.0  \\
    \midrule
    + {\tool} & +3.1 & +0.9 & +1.1 & +0.0 & +0.1 \\
    \bottomrule
    \end{tabular}
    }
    \label{tab:other_generative_tasks}
    \end{sc}
    \end{small}
    \end{center}
    \vskip -0.1in
\end{table}

While we focus on code generation in this work, we are interested in whether  {\tool} can be applied to other generative tasks, Thus, we evaluate {\tool} on other generative benchmarks, including TruthfulQA~\cite{truthfulqa}, GSM8K~\cite{gsm8k}, MMLU~\cite{mmlu}, and BoolQ~\cite{boolq}.
As shown in Table~\ref{tab:other_generative_tasks}, {\tool} enhances the base model's performance in all benchmarks except MMLU. 
While the performance improvement is not as significant as in code generation, we hypothesize that this is because different tasks have unique input and output patterns, leading to varying degrees of attention dilution.
For MMLU, the model only needs to generate a choice to answer the multiple-choice question. The output length is significantly shorter than in code generation and considerably less than in user prompts. Consequently, the model's attention is hardly diluted by self-generated tokens, making {\tool} not helpful.
In contrast, TruthfulQA requires the model to generate a text analysis, provide reasoning, and answer the question. 
Therefore, {\tool} is more beneficial in addressing attention dilution and correcting the remaining 27\% errors for ``\textsc{Truth.}'' (Truthfulness) and 41\% errors for ``\textsc{Info.}'' (Informativeness). 
Nevertheless, it remains an interesting future work to investigate how to further improve {\tool} for a broader range of generative tasks beyond code generation. 
More details are reported in Appendix~\ref{app:other_generative_tasks}.

\section{Related Work}

\textbf{Code Generation.}
To enhance the performance of LLMs on coding tasks, significant efforts have been dedicated to curating high-quality training data~\citep{starcoder1, deepseek_coder, sql_domain_adaptation, wei2023magicoder} and designing domain-specific training objectives~\citep{niu2022spt, chakraborty2022natgen}. Furthermore, techniques such as instruction tuning~\citep{instruction_tuning}, reinforcement learning with human feedback~\citep{rlhf}, and repository-level context modeling~\citep{repo_coder} have been explored to improve alignment, reasoning, and context understanding abilities in code generation. 
However, these approaches often require significant fine-tuning effort. In the meantime, a line of work has focused on developing interactive methods~\citep{di2025, steps_emnlp, sqlucid} that incorporate real-time human feedback to guide and refine the model’s output. Despite their effectiveness, such human-in-the-loop approaches remain limited due to the requirement of user intervention. 
To overcome these challenges, recent research has increasingly explored automated and training-free prompting methods~\citep{self_debug, self_edit, cot_code1, codechain}, which aim to automatically optimize prompts by integrating additional reasoning or contextual information.
For example, Self-Debugging \citep{self_debug} enable LLMs to debug code based on error messages and execution results.
Self-Planning \citep{self_plan} allows LLMs to decompose tasks into subtasks and implement solutions step-by-step. 
ReAct~\cite{react} prompts an LLM to generate reasoning traces and action plans in an interleaved manner.
Compared with these methods, {\tool} takes an orthogonal approach that amplifies the influence of the user prompt to mitigate the attention dilution issue.

\textbf{Attention Steering.}
TOAST~\cite{toast} tunes a feature selection module to redirect 
attention to task-relevant features.
PASTA~\cite{pasta} performs model profiling to identify beneficial attention headers and recalculate attention distributions across transformer layers.
However, these attention-steering approaches usually require extensive model adaptations and complex setup procedures.
In contrast, {\tool} provides a model-agnostic solution that 
mathematically simulates attention steering via logit arithmetic.

\textbf{Logit Arithmetic.}
There has been a growing body of research on performing arithmetic transformations on logits to enhance text generation, such as contrasting logits from multiple LMs \citep{proxy_tuning, dexperts, domain_differential_adaptation, weaktostrong} and contrasting logits from different layers of a model \citep{dola, autocontrastive_decoding}.
Unlike these methods, {\tool} contrasts logits from the same model by perturbing the input through masking, rather than providing additional context~\citep{prefix_adaptive_decoding, context_aware_decoding, coherence} or changing to a completely new prompt~\citep{mitigating_Hallucinations_translation}.
Furthermore, we provide a mathematical approximation of semantic scaling over arbitrary groups of embeddings. 
{\tool} is specifically designed to address the attention dilution issue in LLMs during code generation---a phenomenon first observed in our work.
By contrast, none of the existing works explored code generation or model attention.
They primarily focus on enhancing coherence \citep{coherence}, factuality~\citep{context_aware_decoding, dola, mitigating_Hallucinations_translation, mitigating_Hallucinations_vision}, and controllability~\citep{dexperts, prefix_adaptive_decoding, weaktostrong}.

\section{Limitations \& Future Work}
\label{sec:limitation}


In this work, we pre-define the method for selecting anchored tokens and use a fixed anchoring strength when generating code. We consider this approach a baseline. 
Future work could explore how to dynamically select the \textit{anchored text} and \textit{anchoring strength}.
 For the selection of \textit{anchored text}, one idea is to use LLMs to dynamically identify relevant words or phrases corresponding to the current generation step. 
For tasks where NL may not be important compared to code (e.g., code translation), we can use static code analysis to identify important code elements (e.g., function calls and variable names heavily used in the code).
For the selection of \textit{anchoring strength}, one idea is to develop a method to calculate the relevance of words and phrases to each generation step. Based on the relevance scores, the system can assign higher values to more relevant contexts while assigning lower values to less relevant ones.

In our current experiments, we utilize existing test cases in the benchmark to determine whether to activate {\tool}. Although this is a common practice in code generation~\cite{self_debug, self_edit}, this may not be applicable to programming tasks where test cases are not available. This can be potentially addressed by prompting LLMs to generate test cases.
Furthermore, we believe {\tool} will be beneficial in a self-improving pipeline for fairly complex tasks. By analyzing the errors in the initially generated code, LLMs can be prompted to identify which instructions or requirements were not followed in the prompt. Then {\tool} can be used to amplify the influence of these ignored instructions.


\section{Conclusion}
We present {\tool}, a model-agnostic approach to enhancing LLM code generation. Our study first identifies the attention dilution phenomenon, where code LLMs increasingly overlook the prompt as generation progresses. To address this, {\tool} introduces a training-free, mathematically proven mechanism for controlling the influence of selected prompt tokens. 
Experiments demonstrate that aligning model attention to user prompts using {\tool} significantly and consistently enhances the code generation performance of base LLMs.

\section*{Acknowledgements}
We thank all the anonymous reviewers for their valuable and detailed feedback, which has significantly improved the quality of this paper. This work was supported in part by the National Science Foundation (NSF CAREER Grant 2340408).



\section*{Impact Statement}

This paper presents work whose goal is to advance the field of 
Machine Learning. There are many potential societal consequences 
of our work, none of which we feel must be specifically highlighted here.


\bibliography{custom}
\bibliographystyle{icml2025}

\newpage
\appendix
\onecolumn
\appendix


\onecolumn

\section*{Table of Contents}

\hspace*{1em} A\quad Performance Analysis by Different Generation Lengths \dotfill \pageref{app:generation_length_analysis} \\[0.1cm]
\hspace*{1em} B\quad Calculation Details and Extended Discussion of LLM Attention \dotfill \pageref{app:attention_calculation} \\[0.1cm]
\hspace*{2em} B.1\quad Calculation of Self-attention \dotfill 
\pageref{app_sub:calculation_self_attention} \\[0.1cm]
\hspace*{2em} B.2\quad Calculation of Gradient-based Attention \dotfill 
\pageref{app_sub:calculation_gradient_attention} \\[0.1cm]
\hspace*{2em} B.3\quad Attention Misalignment \dotfill 
\pageref{app_sub:attention_misalignment} \\[0.1cm]
\hspace*{2em} B.4\quad Attention to the User Prompt Ratio \dotfill 
\pageref{app_sub:attention_ratio} \\[0.1cm]
\hspace*{1em} C\quad Extended Discussion of Approximation in {\tool} \dotfill \pageref{app:Higher_order_approximation} \\[0.1cm]
\hspace*{1em} D\quad Code Generation Examples \dotfill \pageref{app:examples} \\[0.1cm]
\hspace*{1em} E\quad Computational Cost of {\tool} \dotfill \pageref{app:computation} \\[0.1cm]
\hspace*{2em} E.1\quad Latency \dotfill 
\pageref{app_sub:latency} \\[0.1cm]
\hspace*{2em} E.2\quad Memory \dotfill 
\pageref{app_sub:memory} \\[0.1cm]
\hspace*{2em} E.3\quad Tradeoff Between Latency and Memory Usage via Parallelization \dotfill 
\pageref{app_sub:computation_tradeoff} \\[0.1cm]
\hspace*{1em} F\quad Pass@10 \& Beam Search with {\tool} \dotfill \pageref{app:beam_search} \\[0.1cm]
\hspace*{1em} G\quad Hypothetical explanation for Attention dilution and SPA's effectiveness \dotfill \pageref{app:explanation}
\\[0.1cm]
\hspace*{1em} H\quad Implementation and Deployment \dotfill \pageref{app:model_set_up} \\[0.1cm]
\hspace*{2em} H.1\quad Imlementation of {\tool} \dotfill 
\pageref{subapp:implementation} \\[0.1cm]
\hspace*{2em} H.2\quad Model Deployment \dotfill 
\pageref{subapp:model_deployment} \\[0.1cm]
\hspace*{2em} H.3\quad Prompt Design \dotfill 
\pageref{subapp:prompt} \\[0.1cm]
\hspace*{1em} I\quad Evaluating {\tool} on Other Generative Tasks \dotfill \pageref{app:other_generative_tasks} \\[0.1cm]
\hspace*{1em} J\quad Tuning of Anchoring Strength \dotfill \pageref{app:tuning_study} \\[0.1cm]
\hspace*{2em} J.1\quad Tuning Stability \dotfill \pageref{app_sub:tuning_stability} \\[0.1cm]
\hspace*{2em} J.2\quad Cross-dataset \& Cross-Model Tuning Evaluation \dotfill \pageref{app_sub:cross_tuning} \\[0.1cm]
\hspace*{2em} J.3\quad Comparision between Tuned {\tool} and Optimal {\tool} \dotfill \pageref{app_sub:optimal_and_tuned} \\[0.1cm]
\hspace*{1em} K\quad Analysis of {\tool}'s Performance without Test Case Availability \dotfill \pageref{app:no_test_case} \\[0.1cm]
\hspace*{1em} L\quad Comparison to \textsc{Pasta} \dotfill 
\pageref{app:comparison_to_pasta} \\[0.1cm]
\hspace*{1em} M\quad Confidence-Modulated Anchoring Strength \dotfill 
\pageref{app:confidence_modulate} \\[0.1cm]


\section{Performance Analysis by Different Generation Lengths}
\label{app:generation_length_analysis}

\begin{figure}[!h]
    \centering
    \includegraphics[width=\linewidth]{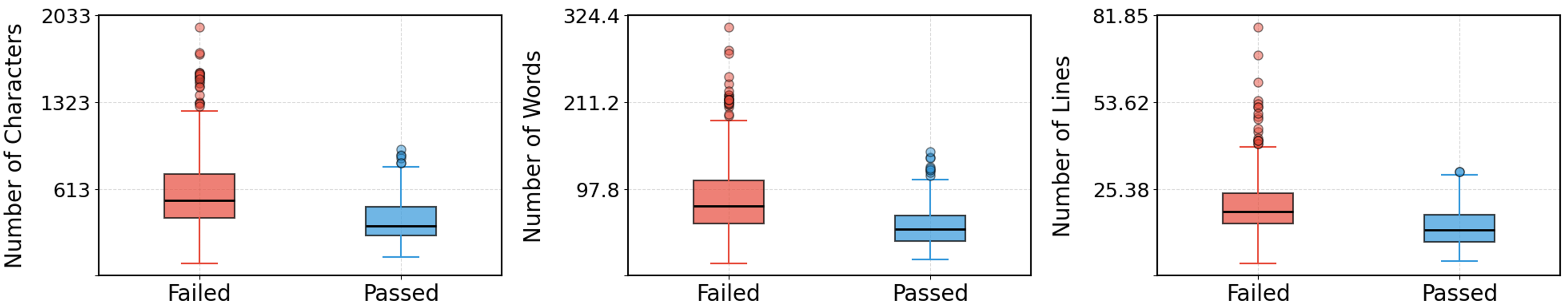}
    \caption{Length of correct vs.~incorrect code generated by LLMs at three levels of granularity: character, word, and line.}
    \label{fig:generation_length_app}
\end{figure}

To evaluate how code length affects model performance, we analyze the length distribution of correctly and incorrectly generated code.
We use the same set of experimental LLMs~\cite{codegen, deepseek_coder, code_llama} as in Section~\ref{sec:empirical} and experiment on both HumanEval~\cite{humaneval} and LiveCodeBench~\cite{livecodebench}.

Particularly, we evaluate code length at three granularities: characters, words (tokens), and lines.
Figure~\ref{fig:generation_length_app} demonstrates that incorrectly generated code is significantly longer than correctly generated code across different granularity levels.
To mitigate the impact of the correlation between task difficulty and generation length, we conduct further analysis at different task difficulty levels in LiveCodeBench. Table~\ref{tab:generation_length} shows that the average length of incorrectly generated code is consistently longer than that of correctly generated code, across all levels.

These results echo our findings about attention dilution---as the generated sequence grows longer, the model's attention to the user prompt diminishes, leading to more errors.





\section{Calculation Details and Extended Discussion of LLM Attention}
\label{app:attention_calculation}

\subsection{Calculation of Self-attention}
\label{app_sub:calculation_self_attention}

Most LLMs are based on the decoder of transformer~\citep{attention_is_all_you_need}, which has multiple self-attention layers.
Roughly speaking, given an LLM \( {\model} \) and an input sequence of tokens \( t_0, t_1, \ldots, t_n \) where $t_i$ represents the $i$th token.
The transformer calculates relevance scores between every pair of tokens. The self-attention score for a token \( t_i \) in the sequence can be roughly formulated as:
\begin{equation}
\mathrm{attention}(t_i) \approx \frac{\sum_{j=1}^n \mathrm{relevance}(t_i, t_j)}{\sum_{i=1}^n \sum_{j=1}^n \mathrm{relevance}(t_i, t_j)},
\end{equation}
where the relevance function approximates the computation among $Q, K, V$ in transformers~\citep{attention_is_all_you_need}.
However, different layers have different attention distributions.
According to a study~\citep{wan2022capture}, deeper self-attention layers can better capture long-distance dependencies and program structure, so we calculate the attention by aggregating attention from multiple heads at the last layer.
Nevertheless, this still excludes the influence from the last forward layer.

\subsection{Calculation of Gradient-based Attention}
\label{app_sub:calculation_gradient_attention}

To validate the generalizability of attention dilution, we employ a gradient-based attention calculation method~\citep{gradient1, gradient2}. 
Compared to using self-attention layers in transformers, the gradient-based method can be generalized to different model architectures by treating the entire model as a whole differentiable function.
It computes the model's attention by calculating the gradients relative to each input token.
Intuitively, a token that induces a larger gradient is considered more influential, suggesting that the model pays greater attention to it. 
Formally, the attention over the token $t_i$ is calculated by
\begin{equation}
    \mathrm{attention}(t_i) = \frac{\partial {\model}(t_0, t_1, \ldots, t_n) }{\partial t_i}.
\end{equation}

\begin{figure}[!h]
    \centering
    \includegraphics[width=0.63\linewidth]{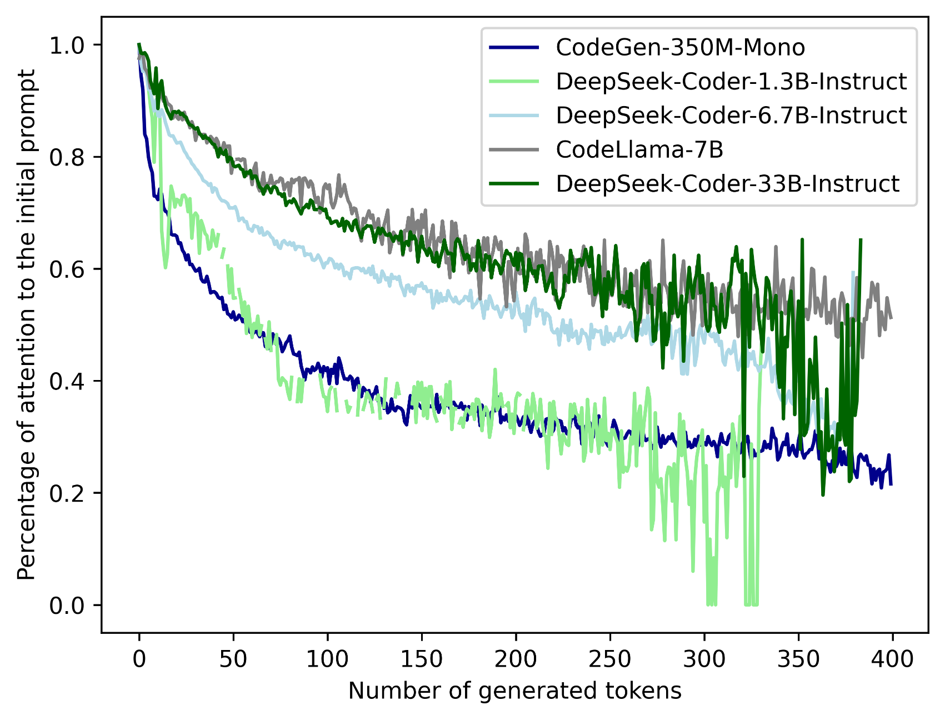}
    \caption{Shift of LLMs' gradient-based attention to the initial prompt. The gradient is calculated with respect to the output logits.}
    \label{fig:gradient_attention}
\end{figure}

As shown in Figure~\ref{fig:gradient_attention}, we observe a similar declining pattern in the model's attention over the initial prompt, suggesting that attention dilution is a fundamental phenomenon that persists across different attention measurement approaches.

\subsection{Attention Misalignment}
\label{app_sub:attention_misalignment}
Despite the success of the attention mechanism in LLMs, prior works found that language models often exhibit simple attention patterns~\citep{simple_attention1, simple_attention2}.
Furthermore, an empirical study~\citep{model_human_attention_align} found that given a coding task, there often exists a misalignment between LLM attention and human attention. 
When generating code, LLMs often focus on parts in natural language descriptions that are different from what human programmers focus on.
Another work~\citep{tiis} shows that when the model's attention aligns more closely with human programmers' attention, the model generates more accurate SQL queries.
Inspired by these findings, we hypothesize that a root cause of inaccuracy in LLM-generated code stems from the suboptimal model attention.

\subsection{Attention to the User Prompt Ratio}
\label{app_sub:attention_ratio}
Based on these two methods to calculate LLMs' attention, we analyze how the attention of LLMs to the initial prompt shifts.
Formally, given the prompt $x$ and the following generated tokens $t_0, t_1, \ldots, t_{i-1}$, we calculate the \textit{attention to the user prompt ratio} $\alpha (x)$ over the initial prompt 
\begin{equation}
    \alpha (x) = \frac{\mathrm{attention}(x)}{\mathrm{attention}(x) + 
      \sum^{n}_{i=1} \mathrm{attention}(t_i)}
\end{equation}

Given that attention analysis requires open sourcing, we select five SOTA code LLMs with various sizes.
We run the experiments on HumanEval~\citep{humaneval}, one of the most popular benchmarks for evaluating code generation models.
We run five LLMs~\citep{codegen, code_llama, deepseek_coder} on all 164 Humaneval tasks.
Figure~\ref{fig:gradient_attention} and Figure~\ref{fig:self_attention} show the gradient-based attention shift when generating the first 400 tokens.
The value gradually becomes noisy due to the insufficient length of the generated sequence.

The results demonstrate that there indeed exists such an attention dilution issue.
Due to the autoregressive nature, LLMs' attention to the initial prompt is gradually diluted as they generate more code. LLMs tend to attend to code generated by themselves. 
Our finding is supported by another study~\citep{attention_dilute} which investigates the self-attention dilution of transformers in a more general scenario.

\section{Extended Discussion of Approximation in {\tool}}
\label{app:Higher_order_approximation}

In Section~\ref{sec:augmented_logits_approximation}, Equation~\ref{eq:12} delivers the approximation by only keeping the first derivative in Equation~\ref{eq:taylor}, but it is also feasible to calculate a higher-order approximation.
For example, if we want to keep the term involving the second-order derivative $\frac{{\weight}^2}{2!}{\logits}''(0)$, it can still be computed using finite-difference methods:
\begin{equation}
{\logits}''(0) \approx \frac{{\logits}(1)-2{\logits}(0)+{\logits}(-1)}{(1-0)^2}.
\end{equation}
${\logits}(-1)$ can be solved by Equation~\ref{eq:14} where ${\logits}(0)$ and ${\logits}(1)$ are the logits generated from the original input and the logits generated from the masked input.

However, no matter how many terms we keep in Equation~\ref{eq:taylor}, we find we can only represent ${\logits}({\weight})$ as a linear combination of $F(0)$ and $F(1)$, weighted by an unknown variable ${\weight}$.

In Section~\ref{sec:strength_impact}, our experiments reveal that ${\weight}$'s impact on code generation performance follows an unimodal pattern—initially increasing, then decreasing. Due to its distribution simplicity, we argue that while a higher-order approximation may yield a more reasonable performance distribution across different ${\weight}$ values, it does not significantly affect the process of locating the optimal anchoring strength.
Therefore, beyond its computational efficiency, the first-order approximation in {\tool} is adequate for calculating semantically accurate augmented logits.

\section{Code Generation Examples}
\label{app:examples}
\begin{figure*}[!h]
    \centering
    \includegraphics[width=0.8\linewidth]{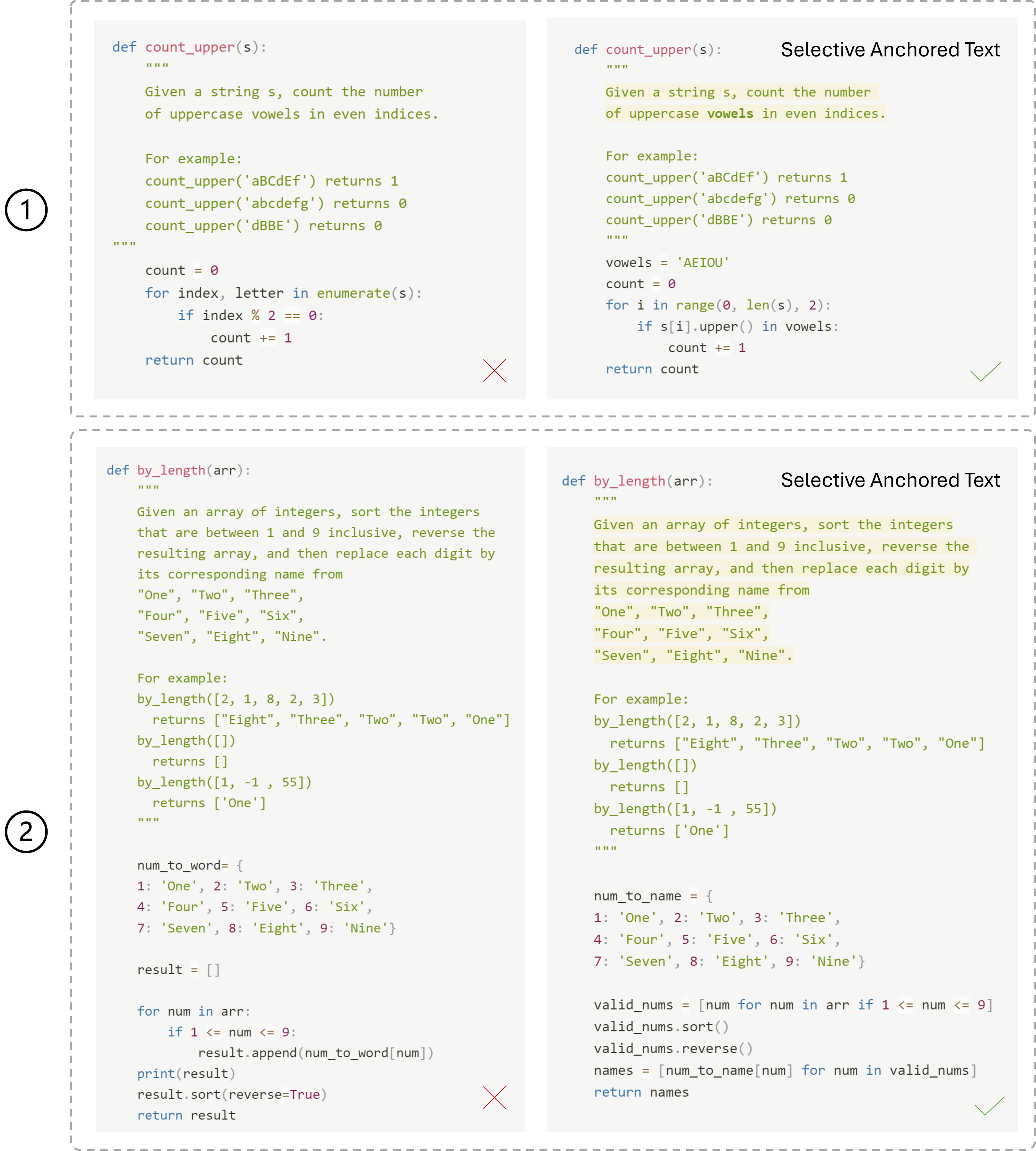}
    \caption{Examples of generated code by LLMs alone (left) and using {\tool} (right).}
    \label{fig:example}
\end{figure*}

Figure~\ref{fig:example} presents two examples comparing the code generated by models alone and the models augmented using {\tool}.

In the first example, CodeLlama (7B) overlooks the specified condition "upper vowels." In contrast, {\tool} enhances the model's focus on the intended purpose. The code initializes all the upper vowels in the first line and correctly refers to it later.

In the second example, DeepSeek-Coder (1.3B) erroneously sorts the list by string names instead of integers. When using {\tool}, the model demonstrates improved recognition of the required procedures, aligning more closely with the task description. The code correctly sorts and reverses the list. Then the integer list is mapped to the string list.

\section{Computational Cost of {\tool}}
\label{app:computation}

\subsection{Latency}
\label{app_sub:latency}
We analyzed the sources of inference overhead, which include running test cases, the two decoding processes, and the logit arithmetic operations. 
We observed that each attention augmentation in {\tool} results in an additional 0.6 times the base model's latency, but this increase occurs only when test cases fail.
Running test cases incurs only a very small cost (0.1s on average) compared to the total inference time (9.6s). 
Thus, only activating {\tool} when test failures are detected can reduce the overall overhead. The more accurate the base model, the less overhead {\tool} introduces.
On average, {\tool} increases the inference time to 1.27 times that of the original model.
We believe this overhead is acceptable in practical development.


\begin{table}[h]
    \caption{Comparison of Tokens Per Second With and Without {\tool}.}
    \label{tab:speed_app}
    \vskip 0.15in
    \begin{center}
    \begin{small}
    \begin{sc}
    \begin{tabular}{lc}
    \toprule
    Model & Token/Second \\
    \midrule
    Codegen (350M) & 34.1 \\
    +{\tool} & 22.7 \\
    \midrule
    DeepSeek-Coder (1.3B) & 17.8 \\
    +{\tool} & 14.4 \\
    \midrule
    DeepSeek-Coder (6.7B) & 12.1 \\
    +{\tool} & 10.2 \\
    \midrule
    CodeLlama (7B) & 14.5 \\
    +{\tool} & 10.6 \\
    \midrule
    StarCoder (15B) & 7.4 \\
    +{\tool} & 6.2 \\
    \midrule
    DeepSeek-Coder (33B) & 5.3 \\
    +{\tool} & 4.6 \\
    \bottomrule
    \end{tabular}
    \end{sc}
    \end{small}
    \end{center}
    \vskip -0.1in
\end{table}

\subsection{Memory}
\label{app_sub:memory}

SPA requires the storage of the logit generated by the masked prompt embeddings. Theoretically, the additional memory requirement, denoted as $M_{overhead}$, can be expressed as:
\begin{equation}
M_{overhead} = V \times D \times S_{logit}
\end{equation}
where $V$ represents the vocabulary size, $D$ is the token embedding dimension, and $S_{logit}$ is the size of a single logit value.

Consider a vocabulary size $V = 50,000$, a token embedding dimension $D = 4,096$, and a logit size $S_{logit} = 2 \text{ bytes}$. Then the additional memory overhead is calculated as:
\begin{equation}
50,000 \times 4,096 \times 2 \text{ bytes} \approx 390 \text{ MB}
\end{equation}
In practice, this memory overhead can be significantly reduced because many low-ranked tokens hardly contribute to the results. We can calculate the augmented logits by considering only the top-ranked tokens and ignoring the rest. For example, if we focus only on the top 100 logits, the overhead will dramatically decrease to $800 \text{ KB}$.


\subsection{Tradeoff Between Latency and Memory Usage via Parallelization}
\label{app_sub:computation_tradeoff}
Currently, {\tool} is implemented by sequentially computing the logits. However, the latency can be further reduced by parallelizing the logits generation from the original and masked embeddings, as these computations are independent (Figure~\ref{fig:pipeline}). This optimization, however, comes at the cost of increased memory usage (double VRAM or memory for forwarding). Given that LLM inference is memory-intensive, this introduces a tradeoff between reduced latency and higher memory consumption.

\section{Pass@10 \& Beam Search with {\tool}}
\label{app:beam_search}

To further evaluate the generalizability of {\tool}, we assess its Pass@10 performance using beam search.
While {\tool} produces augmented logits that could be used directly for beam search, we observed that directly sampling top beams from these augmented logits does not improve the performance.
We hypothesize that this phenomenon occurs because while {\tool} successfully amplifies the influence of anchored text and improves the accuracy of top logits, it also amplifies noise in lower-ranked logits. This undermines the reliability of the overall probability distribution, thereby hindering the sampling process. 

To address this issue, we retrieve top candidate tokens based on the augmented logits but use original probabilities to accumulate beam probability. This strategy ensures that important, potentially overlooked tokens are considered while maintaining reliable probabilities.

\begin{table*}[h!]
    \centering
    \caption{Pass@$1$ and Pass@$10$ (\%) with and without using {\tool}.}
    \label{tab:pass10}     
    \setlength{\tabcolsep}{2pt} 
    \vskip 0.15in
    \begin{center}
    \begin{small}
    \begin{sc}
    \begin{tabular}{
        l
        l
        ll
        ll
        ll
        ll
    }
        \toprule
        \multirow{2}{*}[-0.5ex]{{Model}} & \multirow{2}{*}[-0.5ex]{{Size}} & 
        \multicolumn{2}{c}{{HumanEval}} & 
        \multicolumn{2}{c}{{HumanEval+}} & 
        \multicolumn{2}{c}{{MBPP}} & 
        \multicolumn{2}{c}{{MBPP+}} \\
        \cmidrule(lr){3-4} \cmidrule(lr){5-6} \cmidrule(lr){7-8} \cmidrule(lr){9-10}
        & & \text{\footnotesize     Pass@1} & \text{\footnotesize     Pass@10} & \text{\footnotesize     Pass@1} & \text{\footnotesize     Pass@10} & \text{\footnotesize     Pass@1} & \text{\footnotesize     Pass@10} & \text{\footnotesize     Pass@1} & \text{\footnotesize     Pass@10} \\
        \midrule
        \text{\small CodeGen-Mono} & \smallerfootnote{(350M)} & 15.3 & 36.6 & 12.2 & 33.6 & 19.6 & 47.7 & 15.9 & 42.4 \\
                + {\tool} &  & 20.1 \textcolor{mygreen}{\tiny(+4.8)} & 39.0 \textcolor{mygreen}{\tiny(+2.4)} & 17.1 \textcolor{mygreen}{\tiny(+4.9)} & 38.5 \textcolor{mygreen}{\tiny(+4.9)} & 27.4 \textcolor{mygreen}{\tiny(+7.8)} & 55.6 \textcolor{mygreen}{\tiny(+7.9)} & 22.6 \textcolor{mygreen}{\tiny(+6.7)} & 43.3 \textcolor{mygreen}{\tiny(+0.9)} \\
        \midrule
        \text{\small DeepSeek-Coder} & \smallerfootnote{(1.3B)} & 66.4 & 73.3 & 61.8 & 68.7 & 58.2 & 67.0 & 52.4 & 63.7 \\
                + {\tool} &  & 70.1 \textcolor{mygreen}{\tiny(+3.7)} & 73.3 \textcolor{lightgray}{\tiny(+0.0)} & 67.7 \textcolor{mygreen}{\tiny(+5.9)} & 69.3 \textcolor{mygreen}{\tiny(+0.6)} & 60.9 \textcolor{mygreen}{\tiny(+2.7)} & 68.8 \textcolor{mygreen}{\tiny(+1.8)} & 52.4 \textcolor{lightgray}{\tiny(+0.0)} & 64.3 \textcolor{mygreen}{\tiny(+0.6)} \\
        \midrule
        \text{\small DeepSeek-Coder} & \smallerfootnote{(6.7B)} & 75.6 & 84.0 & 70.2& 77.9 & 67.0 & 79.8 & 58.5 & 70.2 \\
                + {\tool} &  & 88.5 \textcolor{mygreen}{\tiny(+12.9)} & 86.4 \textcolor{mygreen}{\tiny(+2.4)} & 79.9 \textcolor{mygreen}{\tiny(+9.7)} & 82.8 \textcolor{mygreen}{\tiny(+4.9)} & 71.0 \textcolor{mygreen}{\tiny(+4.0)} & 87.7 \textcolor{mygreen}{\tiny(+7.9)} & 60.7 \textcolor{mygreen}{\tiny(+2.2)} & 73.9 \textcolor{mygreen}{\tiny(+3.7)} \\
        \midrule
        \text{\small CodeLlama} & \smallerfootnote{(7B)} & 33.6 & 58.0 & 28.2 & 48.9 & 50.9 & 61.0 & 40.8 & 49.0 \\
                + {\tool} &  & 44.0 \textcolor{mygreen}{\tiny(+10.4)} & 64.7 \textcolor{mygreen}{\tiny(+6.7)} & 36.0 \textcolor{mygreen}{\tiny(+7.8)} & 54.4 \textcolor{mygreen}{\tiny(+5.5)} & 54.3 \textcolor{mygreen}{\tiny(+3.4)} & 65.3 \textcolor{mygreen}{\tiny(+4.3)} & 44.0 \textcolor{mygreen}{\tiny(+3.2)} & 52.0 \textcolor{mygreen}{\tiny(+3.0)} \\
        \midrule
        \text{\small DeepSeek-Coder} & \smallerfootnote{(33B)} & 81.7 & 88.5 & 77.1 & 80.2 & 73.4 & 86.8 & 63.2 & 75.8 \\
                + {\tool} & & 86.2 \textcolor{mygreen}{\tiny(+4.5)} & 89.7 \textcolor{mygreen}{\tiny(+1.2)} & 79.3 \textcolor{mygreen}{\tiny(+2.2)} & 83.2 \textcolor{mygreen}{\tiny(+3.0)} & 79.4 \textcolor{mygreen}{\tiny(+6.0)} & 89.2 \textcolor{mygreen}{\tiny(+2.4)} & 70.3 \textcolor{mygreen}{\tiny(+7.1)} & 80.1 \textcolor{mygreen}{\tiny(+4.3)} \\
        \bottomrule
    \end{tabular}
    \end{sc}
    \end{small}
    \end{center}
    \vskip -0.1in
\end{table*}

As shown in Table~\ref{tab:pass10}, {\tool} consistently improves Pass@10 by, on average, 3.42\% and up to 7.9\% when using beam search.
While the improvements are not as pronounced as those seen with Pass@1, we anticipate that future work could develop beam search algorithms specifically optimized for {\tool}'s unique logit distribution characteristics.

\section{Hypothetical explanation for Attention dilution and {\tool}'s effectiveness} 
\label{app:explanation}

{\tool} is motivated by a recent study~\cite{model_human_attention_align} and our empirical observations demonstrating the attention dilution issue. 
Our experiment results in Section~\ref{sec:results} echo our observation and confirm the existence of attention dilution during code generation.
Here we propose a detailed explanation for this phenomenon based on our knowledge and hypotheses. We believe it stems from two limitations in regular autoregressive decoding: (1) \textbf{Distraction} and (2) \textbf{Error propagation}.

\noindent \textbf{Distraction.} When a transformer generates a token, its correctness depends on two abilities: (1) whether the model attends to the correct context, and (2) whether the model can derive the correct token based on this context. {\tool} aims to improve the first ability. Suppose we have a perfect transformer. For each generated token, it should only attend to relevant prior tokens and ignore irrelevant ones. However, no model is perfect. For each prior token, there is a chance the model incorrectly identifies and attends to it. 
As the model generates more tokens that compete for attention, it becomes increasingly challenging for the model to accurately distribute its attention. The model has more chances to attend to irrelevant tokens, making its attention increasingly unreliable.

In contrast, the user prompt is persistently relevant throughout the generation since it represents the user's intent.
While self-generated tokens are also important context, they are less persistently related than task descriptions in code generation. 
Amplifying the impact of the task description by {\tool} essentially enhances attention reliability, thereby mitigating distraction.

\noindent \textbf{Error propagation.} 
During code generation, the model may generate irrelevant code tokens.
However, the autoregressive nature of LLMs assumes that all previously generated tokens are correct.
For example, if the model introduces an irrelevant variable declaration, subsequent generations may take it into account and continue to generate irrelevant code. 
Although the model can still generate correct code behavior in later generations, this assumption of correctness makes it difficult to identify errors. 
As a result, the error can propagate and accumulate, leading to a higher probability of errors in later generations.

{\tool} mitigates this issue by reinforcing attention to the user prompt while downplaying reliance on self-generated tokens. This optimizes the attention distribution based on the trustworthiness of different contexts, thereby increasing accuracy.




\section{Implementation and Deployment}
\label{app:model_set_up}

\subsection{Implementation of {\tool}}
\label{subapp:implementation}

{\tool} is a model-agnostic algorithm and our implementation does not rely on specific models.
All six models in our paper are built upon the Huggingface Transformer library, which offers APIs to directly access and edit token embeddings and logits. 
Particularly, the {\tool} generator inherits the Huggingface Transformers generation API.\footnote{\url{https://huggingface.co/docs/transformers/en/main_classes/text_generation}}
We leverage the hook to modify the logit calculation within the original generation pipeline.
The API works for any LLM in the Huggingface model collections \footnote{\url{https://huggingface.co/models}} with native hyperparameters such as \textsc{temperature}.
We have released a PyPI library \footnote{\url{https://pypi.org/project/anchoring/}} for developers to quickly test {\tool}.

\subsection{Model Deployment}
\label{subapp:model_deployment}
We downloaded and deployed LLMs from Huggingface.
To expedite evaluations, we apply 8-bit quantization~\citep{gptq, 8_bit_quant} to all models. Prior studies~\citep{quant_effect1, quant_effect2} have demonstrated that this approach has very little impact on LLM performance. We set the \textit{Temperature} to $0$ and the \textit{beam} to $1$ for greedy decoding in all experiments, except for the one described in Appendix~\ref{app:beam_search}.
All experiments were conducted on a 64-bit Ubuntu 22.04 LTS system, equipped with an AMD EPYC 7313 CPU, eight NVIDIA A5500 GPUs, and 512GB of memory. The experiments ran for approximately seven weeks.

\subsection{Prompt Design}
\label{subapp:prompt}
We use the original task descriptions from the datasets as prompts for the text-completion models, CodeLlama and CodeGen-Mono. 
For the remaining models, we format the prompts using the official chat template from HuggingFace.
All experiments are conducted in a zero-shot setting.

\section{Evaluating {\tool} on Other Generative Tasks}
\label{app:other_generative_tasks}

To evaluate the generalizability of {\tool} beyond code generation, we experiment {\tool} on other generative tasks, including TruthfulQA~\cite{truthfulqa}, GSM8K~\cite{gsm8k}, MMLU~\cite{mmlu}, and BoolQ~\cite{boolq}.

TruthfulQA~\cite{truthfulqa} is a benchmark designed to measure models' ability to avoid generating false or misleading information, requiring models to answer questions while remaining truthful.
GSM8K~\cite{gsm8k} tests mathematical reasoning capabilities through grade school math word problems that require multi-step solutions.
MMLU~\cite{mmlu} evaluates models across 57 subjects, including elementary mathematics, US history, computer science, law, and more, comprehensively testing both breadth and depth of knowledge. It provides multiple-choice questions for LLMs to identify the correct answers.
BoolQ~\cite{boolq} consists of naturally occurring yes/no questions from web queries, testing reading comprehension and binary classification abilities.
We use Llama 3.1-Instruct-8B as our base model since the other models in our study are specifically fine-tuned for code tasks.

\begin{table}[h]
    \caption{Evaluating {\tool} on Different Generative Tasks.}
    \vskip 0.15in
    \begin{center}
    \begin{small}
    \begin{sc}
    \begin{tabular}{@{}l@{\hspace{4pt}}c@{\hspace{4pt}}c@{\hspace{4pt}}c@{\hspace{4pt}}c@{\hspace{4pt}}c@{\hspace{4pt}}c@{}}
    \toprule
    \multirow{2}{*}{Model} & TruthfulQA & TruthfulQA & \multirow{2}{*}{GSM8K} & \multirow{2}{*}{MMLU} & \multirow{2}{*}{BoolQ} & \multirow{2}{*}{HumanEval} \\
    & (Truth.) & (Info.) & & & & \\
    \midrule
    Llama-3.1-Instruct-8B & 88.40\% & 97.78\% & 76.5\% & 69.5\% & 83\% & 63.4\% \\
    \midrule
    + {\tool} & + 3.14\% & + 0.91\% & + 1.10\% & + 0\% & + 0.03\% & + \textbf{9.76\%} \\
    \bottomrule
    \end{tabular}
    \label{tab:other_generative_tasks_app}
    \end{sc}
    \end{small}
    \end{center}
    \vskip -0.1in
\end{table}

As shown in Table~\ref{tab:other_generative_tasks}, while {\tool} provides improvements across these tasks, the gains are significantly smaller compared to code generation tasks.
We attribute this performance difference to the unique input-output pattern of different generative tasks.

For MMLU, the model only needs to generate a choice to answer the multiple-choice question. The output length is significantly shorter than in code generation and considerably less than in user prompts. Consequently, the model's attention is hardly diluted by self-generated tokens, making {\tool} not helpful.
In contrast, TruthfulQA requires the model to generate a text analysis, provide reasoning, and answer the question. 
Therefore, {\tool} is more beneficial in addressing attention dilution and correcting the remaining 27\% errors for Truthfulness and 41\% errors for Informativeness. 
In contrast, code generation is typically lengthy and can easily lead to attention dilution. Moreover, code generation prompts serve as persistent instructions, requiring the LLM to maintain focus throughout the generation process. This differs from tasks like translation, where there is no inherent need to consistently anchor attention to specific components of the input.

Therefore, we believe {\tool} is especially suitable for enhancing LLMs' attention for code generation tasks.
Nevertheless, it remains an interesting future work to investigate how to further improve {\tool} for a broader range of generative tasks beyond code generation. 
More details are reported in Appendix~\ref{app:other_generative_tasks}.

\section{Tuning of Anchoring Strength }
\label{app:tuning_study}

As demonstrated in Section~\ref{sec:strength_impact}, different anchoring strengths ${\weight}$ lead to different performance and follow a simple unimodal pattern.
In this section, we report more evaluation of tuning {\tool}.
To ensure fair comparisons, {\tool} is activated for all tasks in all experiments, unlike during inference time where {\tool} is only activated for failed tasks.
Moreover, we tune {\tool} on all the tasks for better generalizability.






\subsection{Tuning Stability}
\label{app_sub:tuning_stability}

We evaluate the stability of tuning anchoring strength by tuning {\tool} on five exclusive subsets of each dataset for each model, as shown in Table~\ref{tab:tuning_stability}.
The average variance across subsets is $0.0046$ for HumanEval/HumanEval+ and $0.0026$ for MBPP/MBPP+, demonstrating the tuning stability.

\begin{table}[h!]
    \centering
    \caption{Tuned Anchoring Strength on Different Subsets.}
    \label{tab:tuning_stability}
    \vskip 0.15in
    \begin{center}
    \begin{small}
    \begin{sc}
    \begin{tabular}{l l c c}
    \toprule
    Model & Subset & HumanEval/+ & MBPP/+ \\
    \midrule
    CodeGen-Mono (350M) & Subset$_1$ & 1.05 & 1.30 \\
    & Subset$_2$ & 1.10 & 1.35 \\
    & Subset$_3$ & 1.20 & 1.25 \\
    & Subset$_4$ & 1.30 & 1.35 \\
    & Subset$_5$ & 1.25 & 1.35 \\
    & Complete & 1.20 & 1.35 \\
    \midrule
    DeepSeek-Coder (1.3B) & Subset$_1$ & 1.05 & 1.20 \\
    & Subset$_2$ & 1.05 & 1.15 \\
    & Subset$_3$ & 1.10 & 1.15 \\
    & Subset$_4$ & 1.00 & 1.20 \\
    & Subset$_5$ & 1.05 & 1.25 \\
    & Complete & 1.05 & 1.20 \\
    \midrule
    DeepSeek-Coder (6.7B) & Subset$_1$ & 1.30 & 1.30 \\
    & Subset$_2$ & 1.25 & 1.20 \\
    & Subset$_3$ & 1.30 & 1.25 \\
    & Subset$_4$ & 1.20 & 1.20 \\
    & Subset$_5$ & 1.35 & 1.25 \\
    & Complete & 1.28 & 1.25 \\
    \midrule
    CodeLlama (7B) & Subset$_1$ & 1.55 & 1.25 \\
    & Subset$_2$ & 1.55 & 1.20 \\
    & Subset$_3$ & 1.50 & 1.20 \\
    & Subset$_4$ & 1.65 & 1.25 \\
    & Subset$_5$ & 1.65 & 1.15 \\
    & Complete & 1.60 & 1.20 \\
    \midrule
    DeepSeek-Coder (33B) & Subset$_1$ & 1.25 & 1.25 \\
    & Subset$_2$ & 1.30 & 1.30 \\
    & Subset$_3$ & 1.40 & 1.30 \\
    & Subset$_4$ & 1.35 & 1.40 \\
    & Subset$_5$ & 1.35 & 1.20 \\
    & Complete & 1.35 & 1.30 \\
    \bottomrule
    \end{tabular}
    \end{sc}
    \end{small}
    \end{center}
    \vskip -0.1in
\end{table}

\subsection{Cross-dataset \& Cross-model Tuning Evaluation}
\label{app_sub:cross_tuning}

We investigate the transferability of this hyperparameter across different models and datasets.
Firstly, we conduct a \textit{cross-dataset} evaluation between HumanEval/HumanEval+ and MBPP/MBPP+, which have distinct prompt formats. We tune ${\weight}$ on HumanEval+ and evaluate Pass@$1$ on MBPP and MBPP+, and vice versa\footnote{The ``plus'' versions of HumanEval and MBPP share identical prompts with their base counterparts, so we only tune once on the plus version.} (denoted as {\tool}$_{cross-dataset}$).
We calculate average Pass@$1$ improvements on the original and plus versions across all baseline models.
Secondly, we perform a \textit{cross-model} evaluation by tuning ${\weight}$ on one model and evaluating Pass@$1$ on the remaining four. For each model, we compute the average Pass@$1$ improvements across all the other models, for HumanEval/HumanEval+ and MBPP/MBPP+ respectively (denoted as {\tool}$_{cross-model}$).
Similar to Section~\ref{sec:results}, {\tool} represents tuning within the split partial dataset, while {\tool}$^*$ represents tuning within the entire dataset.

\begin{table}[h!]
    \centering
    \caption{Pass@1 improvements (\%) based on cross-dataset tuning.}
    \label{tab:cross_dataset_results}
    \vskip 0.15in
    \begin{center}
    \begin{small}
    \begin{sc}
    \begin{tabular}{
        l
        S[table-format=+2.2, retain-explicit-plus]
        S[table-format=+1.2, retain-explicit-plus]
        S[table-format=+2.2, retain-explicit-plus]
        S[table-format=+2.2, retain-explicit-plus]
    }
        \toprule
        {Dataset} & {{\tool}$_{cross-dataset}$} & {{\tool}$_{cross-model}$} & {{\tool}} & {{\tool}$^*$} \\
        \midrule
        {HumanEval/+} & {+ 2.01} & {- 0.29} & {+ 4.36} & {+ 5.11} \\
        {MBPP/+} & {+ 2.50} & {+ 0.37} & {+ 2.86} & {+ 3.57} \\
        \bottomrule
    \end{tabular}
    \end{sc}
    \end{small}
    \end{center}
    \vskip -0.1in
\end{table}

As shown in Table~\ref{tab:cross_dataset_results}, 
we find \finding{the anchoring strength ${\weight}$ tuned on one model is hardly transferred to another. However, ${\weight}$ tuned on one dataset can be transferred to another with reduced but still effective performance.} 
These observations suggest that the anchoring strength is highly model-dependent and partially task-dependent.

We further investigate whether it is possible to find a universal anchoring strength that works for most scenarios. One potential value is the average of the tuned anchoring strengths across benchmarks for each model. 
We apply this fixed anchoring strength to all settings, denoted as {\tool}-preset. 
As shown in Table~\ref{app_tab:fixed_strength}, although the generation accuracy decreases compared to using a tuned strength, {\tool} with the fixed strength can still outperform the baselines.
It implies that {\tool} is effectively deployable in new scenarios once a reasonable value is set.

\begin{table}[!h]
\caption{Comparison between {\tool} and SOTA methods.}
\label{app_tab:fixed_strength}
\vskip 0.15in
\begin{center}
\begin{small}
\begin{sc}
\resizebox{0.43\columnwidth}{!}{
\begin{tabular}{l@{\hspace{1.5em}}c@{\hspace{1.5em}}c}
\toprule
{Method} & {$\Delta$Pass@1 (\%)} & {Time (Sec)} \\
\midrule
Base Model & 0 & 7.7 \\
\midrule
Pasta & +1.2 & 48.8 \\
Self-Debugging & +4.2 & 27.3 \\
Self-Edit & +1.8 & 26.4 \\
Self-Planning & +3.6 & 21.6 \\
ReAct & +1.3 & 28.8 \\
{\tool}-fixed-strength & +5.0 & \textbf{9.8} \\
{\tool} & \textbf{+7.7} & \textbf{9.8} \\
\bottomrule
\end{tabular}
}
\end{sc}
\end{small}
\end{center}
\vskip -0.1in
\end{table}

\subsection{Comparison between Tuned {\tool} and Optimal {\tool}}
\label{app_sub:optimal_and_tuned}

To better understand the tuning efficiency, we tune {\tool} on each complete dataset to obtain the optimal anchoring strength (denoted as {\tool}$^*$).
As shown in Table~\ref{tab:optimal_vs_tuned}, we find that {\tool}$^*$ achieves 20\% higher performance improvements compared to the tuned {\tool} on average.

\begin{table*}[h!]
    \centering
    \caption{Pass@1 Improvements (\%) Comparision between Tuned {\tool} and Optimal {\tool} ({\tool} is applied to all tasks).}
     \vspace{0.5\baselineskip}
    \label{tab:optimal_vs_tuned}
    \vskip 0.15in
    \begin{center}
    \begin{small}
    \begin{sc}
    \setlength{\tabcolsep}{2pt} 
    \begin{tabular}{
        l
        l
        l
        l
        l
        l
    }
        \toprule
        {Model} & {Size} & 
        \multicolumn{1}{c}{{HumanEval}} & 
        \multicolumn{1}{c}{{HumanEval+}} & 
        \multicolumn{1}{c}{{MBPP}} & 
        \multicolumn{1}{c}{{MBPP+}} \\
        \midrule
        \text{CodeGen-Mono} & \small{(350M)} & 15.3 & 12.2 & 19.6 & 15.9 \\
                + {\tool} &  & 18.3 \textcolor{mygreen}{\footnotesize(+3.0)} & 16.0 \textcolor{mygreen}{\footnotesize(+3.8)} & 24.9 \textcolor{mygreen}{\footnotesize(+5.3)} & 20.6 \textcolor{mygreen}{\footnotesize(+4.7)} \\
                + {\tool}$^*$ &  & 18.3 \textcolor{mygreen}{\footnotesize(+3.0)} & 16.0 \textcolor{mygreen}{\footnotesize(+3.8)} & 24.9 \textcolor{mygreen}{\footnotesize(+5.3)} & 20.6 \textcolor{mygreen}{\footnotesize(+4.7)} \\
        \midrule
        \text{DeepSeek-Coder} & \small{(1.3B)} & 66.4 & 61.8 & 58.2 & 52.4 \\
                + {\tool} &  & 69.5 \textcolor{mygreen}{\footnotesize(+3.1)} & 66.4 \textcolor{mygreen}{\footnotesize(+4.6)} & 59.1 \textcolor{mygreen}{\footnotesize(+0.9)} & 52.4 \textcolor{lightgray}{\footnotesize(+0.0)} \\
                + {\tool}$^*$ &  & 71.0 \textcolor{mygreen}{\footnotesize(+4.6)} & 66.4 \textcolor{mygreen}{\footnotesize(+4.6)} & 61.7 \textcolor{mygreen}{\footnotesize(+3.5)} & 53.4 \textcolor{mygreen}{\footnotesize(+1.0)} \\
        \midrule
        \text{DeepSeek-Coder} & \small{(6.7B)} & 75.6 & 70.2 & 67.0 & 58.5 \\
                + {\tool} &  & 83.2 \textcolor{mygreen}{\footnotesize(+7.6)} & 75.6 \textcolor{mygreen}{\footnotesize(+5.4)} & 69.6 \textcolor{mygreen}{\footnotesize(+2.6)} & 60.2 \textcolor{mygreen}{\footnotesize(+1.7)} \\
                + {\tool}$^*$ &  & 84.0 \textcolor{mygreen}{\footnotesize(+8.4)} & 76.3 \textcolor{mygreen}{\footnotesize(+6.1)} & 72.2 \textcolor{mygreen}{\footnotesize(+5.2)} & 61.1 \textcolor{mygreen}{\footnotesize(+2.6)} \\
        \midrule
        \text{CodeLlama} & \small{(7B)} & 33.6 & 28.2 & 50.9 & 40.8 \\
                + {\tool} &  & 40.5 \textcolor{mygreen}{\footnotesize(+6.9)} & 33.6 \textcolor{mygreen}{\footnotesize(+5.4)} & 52.9 \textcolor{mygreen}{\footnotesize(+2.0)} & 43.1 \textcolor{mygreen}{\footnotesize(+2.3)} \\
                + {\tool}$^*$ &  & 41.2 \textcolor{mygreen}{\footnotesize(+7.6)} & 35.9 \textcolor{mygreen}{\footnotesize(+7.7)} & 52.9 \textcolor{mygreen}{\footnotesize(+2.0)} & 43.1 \textcolor{mygreen}{\footnotesize(+2.3)} \\
        \midrule
        \text{DeepSeek-Coder} & \small{(33B)} & 81.7 & 77.1 & 73.4 & 63.2 \\
                + {\tool} &  & 84.7 \textcolor{mygreen}{\footnotesize(+3.0)} & 77.9 \textcolor{mygreen}{\footnotesize(+0.8)} & 77.2 \textcolor{mygreen}{\footnotesize(+3.8)} & 68.5 \textcolor{mygreen}{\footnotesize(+5.3)} \\
                + {\tool}$^*$ & & 85.5 \textcolor{mygreen}{\footnotesize(+3.8)} & 78.6 \textcolor{mygreen}{\footnotesize(+1.5)} & 77.2 \textcolor{mygreen}{\footnotesize(+3.8)} & 68.5 \textcolor{mygreen}{\footnotesize(+5.3)} \\
        \bottomrule
    \end{tabular}
    \end{sc}
    \end{small}
    \end{center}
    \vskip -0.1in
\end{table*}

Table~\ref{tab:optimal_weight} reports optimal anchoring strength values ${\weight}$.
We observe that the average value of $1.28$ can be used to effectively improve performance across all benchmarks for all LLMs.

\begin{table}[!h]
    \centering
    \caption{Optimal Anchoring Strength (${\weight}$) for each model and benchmark.}
    \label{tab:optimal_weight}
    \vskip 0.15in
    \begin{center}
    \begin{small}
    \begin{sc}
    \begin{tabular}{
        l
        c
        c
        c
        c
        c
    }
        \toprule
        {Model} & {HumanEval} & {HumanEval+} & {MBPP} & {MBPP+}  & \textit{Average} \\
        \specialrule{.6pt}{3pt}{3pt}
        \text{CodeGen-Mono}~\textnormal{(350M)} & 1.20 & 1.20 & 1.35 & 1.35 & 1.28 \\
        \specialrule{.02pt}{3pt}{3pt}
        \text{DeepSeek-Coder}~\textnormal{(1.3B)} & 1.05 & 1.05 & 1.20 & 1.20 & 1.13 \\
        \specialrule{.02pt}{3pt}{3pt}
        \text{DeepSeek-Coder}~\textnormal{(6.7B)} & 1.28 & 1.28 & 1.25 & 1.25 & 1.26 \\
        \specialrule{.02pt}{3pt}{3pt}
        \text{CodeLlama}~\textnormal{(7B)}   & 1.60 & 1.60 & 1.20 & 1.20 & 1.40 \\
        \specialrule{.02pt}{3pt}{3pt}
        \text{DeepSeek-Coder}~\textnormal{(33B)}  & 1.35 & 1.35 & 1.30 & 1.30 & 1.33 \\
        \specialrule{.02pt}{3pt}{3pt}
        \textit{Average}  & 1.30 & 1.30 & 1.33 & 1.33 & 1.28 \\
        \bottomrule
    \end{tabular}
    \end{sc}
    \end{small}
    \end{center}
    \vskip -0.1in
\end{table}

\section{Analysis of {\tool}'s Performance without Test Case Availability}
\label{app:no_test_case}

To analyze {\tool}'s effectiveness in scenarios without available test cases, we conduct additional experiments where {\tool} is applied to all code generation tasks ({\tool}$^{all}$). As shown in Table~\ref{tab:results2}, {\tool}$^{all}$ can still achieve significant improvements across all benchmarks and models. For example, on HumanEval, {\tool}$^{all}$ improves DeepSeek-Coder (6.7B)'s Pass@1 by 7.6\% (from 75.6\% to 83.2\%). 


\begin{table}[!h]
\caption{Comparison to PASTA in terms of Pass@1.}
\label{tab:compare_pasta}
\vskip 0.15in
\begin{center}
\begin{small}
\begin{sc}
\resizebox{0.65\columnwidth}{!}{
\begin{tabular}{l@{\hspace{1.5em}}c@{\hspace{1.5em}}c@{\hspace{1.5em}}c@{\hspace{1.5em}}c@{\hspace{1.5em}}c}
\toprule
& HumanEval & HumanEval+ & Mbpp & Mbpp+ & BigCodeBench \\
\midrule
Pasta & \moresmall +1.22 & \moresmall +1.22 & \moresmall +1.17 & \moresmall +0.94 & \moresmall +0.1 \\
\midrule
\tool & \moresmall +7.32 & \moresmall +4.88 & \moresmall +4.22 & \moresmall +3.51 & \moresmall +0.4 \\
\bottomrule
\end{tabular}
}
\end{sc}
\end{small}
\end{center}
\vskip -0.1in
\end{table}

\begin{table*}[h!]
    \centering
    \caption{Absolute ($\Delta$) and Relative ($\uparrow$) Performance improvements in Pass@$1$ rates (\%). {\tool}$^{all}$ is applied to all tasks, while {\tool}$^{}$ indicates {\tool} is activated when the original generated code fails test cases.}
    \vspace{0.5\baselineskip}
    \label{tab:results2}
    \begin{center}
     \begin{small}
     \begin{sc}
    \begin{tabular}{
        l
        c
        l
        l
        l
        l
        l
    }
        \toprule
        Model & {\footnotesize Size} & {HumanEval} & {HumanEval+} & {MBPP} & {MBPP+} & {BigCodeBench} \\
        \midrule
        CodeGen-Mono & \small (350M) & 15.3 & 12.2 & 19.6 & 15.9 & 1.1\\[-6pt]
        \multicolumn{2}{c}{} & \rule{4em}{0.01pt} & \rule{4em}{0.01pt} & \rule{4em}{0.01pt} & \rule{4em}{0.01pt} & \rule{4em}{0.01pt} \\
        + {\tool}$^{all}$ & & 18.3\!\textcolor{mygreen}{\tiny$\begin{array}{l}\Delta\!+\!3.0\\(20\%\uparrow)\end{array}$} & 16.0\!\textcolor{mygreen}{\tiny$\begin{array}{l}\Delta\!+\!3.8\\(31\%\uparrow)\end{array}$} & 24.9\!\textcolor{mygreen}{\tiny$\begin{array}{l}\Delta\!+\!5.3\\(27\%\uparrow)\end{array}$} & 20.6\!\textcolor{mygreen}{\tiny$\begin{array}{l}\Delta\!+\!4.7\\(30\%\uparrow)\end{array}$} & 1.4\!\textcolor{mygreen}{\tiny$\begin{array}{l}\Delta\!+\!0.3\\(27\%\uparrow)\end{array}$}\\[-6pt]
        \multicolumn{2}{c}{} & \rule{4em}{0.01pt} & \rule{4em}{0.01pt} & \rule{4em}{0.01pt} & \rule{4em}{0.01pt} & \rule{4em}{0.01pt} \\
        + {\tool}$^{}$ & & 20.1\!\textcolor{mygreen}{\tiny$\begin{array}{l}\Delta\!+\!4.8\\(31\%\uparrow)\end{array}$} & 17.1\!\textcolor{mygreen}{\tiny$\begin{array}{l}\Delta\!+\!4.9\\(40\%\uparrow)\end{array}$} & 27.4\!\textcolor{mygreen}{\tiny$\begin{array}{l}\Delta\!+\!7.8\\(40\%\uparrow)\end{array}$} & 22.6\!\textcolor{mygreen}{\tiny$\begin{array}{l}\Delta\!+\!6.7\\(42\%\uparrow)\end{array}$} & 1.6\!\textcolor{mygreen}{\tiny$\begin{array}{l}\Delta\!+\!0.5\\(45\%\uparrow)\end{array}$}\\
        \specialrule{.02pt}{2pt}{2pt}
        DeepSeek-Coder & \small (1.3B) & 66.4 & 61.8 & 58.2 & 52.4 & 2.5\\[-6pt]
        \multicolumn{2}{c}{} & \rule{4em}{0.01pt} & \rule{4em}{0.01pt} & \rule{4em}{0.01pt} & \rule{4em}{0.01pt} & \rule{4em}{0.01pt} \\
        + {\tool}$^{all}$ & & 69.5\!\textcolor{mygreen}{\tiny$\begin{array}{l}\Delta\!+\!3.1\\(5\%\uparrow)\end{array}$} & 66.4\!\textcolor{mygreen}{\tiny$\begin{array}{l}\Delta\!+\!4.6\\(7\%\uparrow)\end{array}$} & 59.1\!\textcolor{mygreen}{\tiny$\begin{array}{l}\Delta\!+\!0.9\\(2\%\uparrow)\end{array}$} & 52.4\!\textcolor{lightgray}{\tiny$\begin{array}{l}\Delta\!+\!0.0\\(0\%\uparrow)\end{array}$} & 3.3\!\textcolor{mygreen}{\tiny$\begin{array}{l}\Delta\!+\!0.8\\(32\%\uparrow)\end{array}$}\\[-6pt]
        \multicolumn{2}{c}{} & \rule{4em}{0.01pt} & \rule{4em}{0.01pt} & \rule{4em}{0.01pt} & \rule{4em}{0.01pt} & \rule{4em}{0.01pt} \\
        + {\tool}$^{}$ & & 70.1\!\textcolor{mygreen}{\tiny$\begin{array}{l}\Delta\!+\!3.7\\(6\%\uparrow)\end{array}$} & 67.7\!\textcolor{mygreen}{\tiny$\begin{array}{l}\Delta\!+\!5.9\\(10\%\uparrow)\end{array}$} & 60.9\!\textcolor{mygreen}{\tiny$\begin{array}{l}\Delta\!+\!2.7\\(5\%\uparrow)\end{array}$} & 52.4\!\textcolor{lightgray}{\tiny$\begin{array}{l}\Delta\!+\!0.0\\(0\%\uparrow)\end{array}$} & 3.4\!\textcolor{mygreen}{\tiny$\begin{array}{l}\Delta\!+\!0.9\\(36\%\uparrow)\end{array}$}\\
        \specialrule{.02pt}{2pt}{2pt}
        {DeepSeek-Coder} & \small (6.7B) & 75.6 & 70.2 & 67.0 & 58.5 & 12.7\\[-6pt]
        \multicolumn{2}{c}{} & \rule{4em}{0.01pt} & \rule{4em}{0.01pt} & \rule{4em}{0.01pt} & \rule{4em}{0.01pt} & \rule{4em}{0.01pt} \\
        + {\tool}$^{all}$ & & 83.2\!\textcolor{mygreen}{\tiny$\begin{array}{l}\Delta\!+\!7.6\\(10\%\uparrow)\end{array}$} & 75.6\!\textcolor{mygreen}{\tiny$\begin{array}{l}\Delta\!+\!5.4\\(8\%\uparrow)\end{array}$} & 69.6\!\textcolor{mygreen}{\tiny$\begin{array}{l}\Delta\!+\!2.6\\(4\%\uparrow)\end{array}$} & 60.2\!\textcolor{mygreen}{\tiny$\begin{array}{l}\Delta\!+\!1.7\\(3\%\uparrow)\end{array}$} & 14.2\!\textcolor{mygreen}{\tiny$\begin{array}{l}\Delta\!+\!1.5\\(12\%\uparrow)\end{array}$}\\[-6pt]
        \multicolumn{2}{c}{} & \rule{4em}{0.01pt} & \rule{4em}{0.01pt} & \rule{4em}{0.01pt} & \rule{4em}{0.01pt} & \rule{4em}{0.01pt} \\
        + {\tool}$^{}$ & & 88.5\!\textcolor{mygreen}{\tiny$\begin{array}{l}\Delta\!+\!12.9\\(17\%\uparrow)\end{array}$} & 79.9\!\textcolor{mygreen}{\tiny$\begin{array}{l}\Delta\!+\!9.7\\(14\%\uparrow)\end{array}$} & 71.0\!\textcolor{mygreen}{\tiny$\begin{array}{l}\Delta\!+\!4.0\\(6\%\uparrow)\end{array}$} & 60.7\!\textcolor{mygreen}{\tiny$\begin{array}{l}\Delta\!+\!2.2\\(4\%\uparrow)\end{array}$} & 16.4\!\textcolor{mygreen}{\tiny$\begin{array}{l}\Delta\!+\!3.7\\(29\%\uparrow)\end{array}$}\\
        \specialrule{.02pt}{2pt}{2pt}
        {CodeLlama} & \small (7B) & 33.6 & 28.2 & 50.9 & 40.8 & 3.4\\[-6pt]
        \multicolumn{2}{c}{} & \rule{4em}{0.01pt} & \rule{4em}{0.01pt} & \rule{4em}{0.01pt} & \rule{4em}{0.01pt} & \rule{4em}{0.01pt} \\
        + {\tool}$^{all}$ & & 40.5\!\textcolor{mygreen}{\tiny$\begin{array}{l}\Delta\!+\!6.9\\(21\%\uparrow)\end{array}$} & 33.6\!\textcolor{mygreen}{\tiny$\begin{array}{l}\Delta\!+\!5.4\\(19\%\uparrow)\end{array}$} & 52.9\!\textcolor{mygreen}{\tiny$\begin{array}{l}\Delta\!+\!2.0\\(4\%\uparrow)\end{array}$} & 43.1\!\textcolor{mygreen}{\tiny$\begin{array}{l}\Delta\!+\!2.3\\(6\%\uparrow)\end{array}$} & 3.8\!\textcolor{mygreen}{\tiny$\begin{array}{l}\Delta\!+\!0.4\\(12\%\uparrow)\end{array}$}\\[-6pt]
        \multicolumn{2}{c}{} & \rule{4em}{0.01pt} & \rule{4em}{0.01pt} & \rule{4em}{0.01pt} & \rule{4em}{0.01pt} & \rule{4em}{0.01pt} \\
        + {\tool}$^{}$ & & 44.0\!\textcolor{mygreen}{\tiny$\begin{array}{l}\Delta\!+\!10.4\\(31\%\uparrow)\end{array}$} & 36.0\!\textcolor{mygreen}{\tiny$\begin{array}{l}\Delta\!+\!7.8\\(28\%\uparrow)\end{array}$} & 54.3\!\textcolor{mygreen}{\tiny$\begin{array}{l}\Delta\!+\!3.4\\(7\%\uparrow)\end{array}$} & 44.0\!\textcolor{mygreen}{\tiny$\begin{array}{l}\Delta\!+\!3.2\\(8\%\uparrow)\end{array}$} & 4.1\!\textcolor{mygreen}{\tiny$\begin{array}{l}\Delta\!+\!0.7\\(21\%\uparrow)\end{array}$}\\
        \specialrule{.02pt}{2pt}{2pt}
        {StarCoder2} & \small (16B) & 67.7 & 60.4 & 78.0 & 65.1 & 13.3\\[-6pt]
        \multicolumn{2}{c}{} & \rule{4em}{0.01pt} & \rule{4em}{0.01pt} & \rule{4em}{0.01pt} & \rule{4em}{0.01pt} & \rule{4em}{0.01pt} \\
        + {\tool}$^{all}$ & & 72.1\!\textcolor{mygreen}{\tiny$\begin{array}{l}\Delta\!+\!4.4\\(6\%\uparrow)\end{array}$} & 63.6\!\textcolor{mygreen}{\tiny$\begin{array}{l}\Delta\!+\!3.2\\(5\%\uparrow)\end{array}$} & 80.9\!\textcolor{mygreen}{\tiny$\begin{array}{l}\Delta\!+\!2.9\\(4\%\uparrow)\end{array}$} & 67.6\!\textcolor{mygreen}{\tiny$\begin{array}{l}\Delta\!+\!2.5\\(4\%\uparrow)\end{array}$} & 14.1\!\textcolor{mygreen}{\tiny$\begin{array}{l}\Delta\!+\!0.8\\(6\%\uparrow)\end{array}$}\\[-6pt]
        \multicolumn{2}{c}{} & \rule{4em}{0.01pt} & \rule{4em}{0.01pt} & \rule{4em}{0.01pt} & \rule{4em}{0.01pt} & \rule{4em}{0.01pt} \\
        + {\tool}$^{}$ & & 75.6\!\textcolor{mygreen}{\tiny$\begin{array}{l}\Delta\!+\!7.9\\(12\%\uparrow)\end{array}$} & 65.6\!\textcolor{mygreen}{\tiny$\begin{array}{l}\Delta\!+\!5.2\\(9\%\uparrow)\end{array}$} & 82.0\!\textcolor{mygreen}{\tiny$\begin{array}{l}\Delta\!+\!4.0\\(5\%\uparrow)\end{array}$} & 69.1\!\textcolor{mygreen}{\tiny$\begin{array}{l}\Delta\!+\!4.0\\(6\%\uparrow)\end{array}$} & 14.3\!\textcolor{mygreen}{\tiny$\begin{array}{l}\Delta\!+\!1.0\\(8\%\uparrow)\end{array}$}\\
        \specialrule{.02pt}{2pt}{2pt}
        {DeepSeek-Coder} & \small (33B) & 81.7 & 77.1 & 73.4 & 63.2 & 18.9\\[-6pt]
        \multicolumn{2}{c}{} & \rule{4em}{0.01pt} & \rule{4em}{0.01pt} & \rule{4em}{0.01pt} & \rule{4em}{0.01pt} & \rule{4em}{0.01pt} \\
        + {\tool}$^{all}$ & & 84.7\!\textcolor{mygreen}{\tiny$\begin{array}{l}\Delta\!+\!3.0\\(4\%\uparrow)\end{array}$} & 77.9\!\textcolor{mygreen}{\tiny$\begin{array}{l}\Delta\!+\!0.8\\(1\%\uparrow)\end{array}$} & 77.2\!\textcolor{mygreen}{\tiny$\begin{array}{l}\Delta\!+\!3.8\\(5\%\uparrow)\end{array}$} & 68.5\!\textcolor{mygreen}{\tiny$\begin{array}{l}\Delta\!+\!5.3\\(8\%\uparrow)\end{array}$} & 20.7\!\textcolor{mygreen}{\tiny$\begin{array}{l}\Delta\!+\!1.8\\(10\%\uparrow)\end{array}$}\\[-6pt]
        \multicolumn{2}{c}{} & \rule{4em}{0.01pt} & \rule{4em}{0.01pt} & \rule{4em}{0.01pt} & \rule{4em}{0.01pt} & \rule{4em}{0.01pt} \\
        + {\tool}$^{}$ & & 86.2\!\textcolor{mygreen}{\tiny$\begin{array}{l}\Delta\!+\!4.5\\(6\%\uparrow)\end{array}$} & 79.3\!\textcolor{mygreen}{\tiny$\begin{array}{l}\Delta\!+\!2.2\\(3\%\uparrow)\end{array}$} & 79.4\!\textcolor{mygreen}{\tiny$\begin{array}{l}\Delta\!+\!6.0\\(8\%\uparrow)\end{array}$} & 70.3\!\textcolor{mygreen}{\tiny$\begin{array}{l}\Delta\!+\!7.1\\(11\%\uparrow)\end{array}$} & 21.5\!\textcolor{mygreen}{\tiny$\begin{array}{l}\Delta\!+\!2.6\\(14\%\uparrow)\end{array}$}\\
        \bottomrule
    \end{tabular}
 \end{sc}
 \end{small}
 \end{center}
 \vskip -0.1in
 \end{table*}

\section{Comparison to \textsc{Pasta}}
\label{app:comparison_to_pasta}

Unlike mainstream methods that optimize prompts, {\tool} enhances performance by optimizing model attention. We compare {\tool} with another attention steering method, PASTA~\cite{pasta}.
PASTA's implementation only supports LLAMA~\cite{llama1}, LLAMA2~\cite{llama2}, and GPT-J~\cite{gptj}. We find that adapting PASTA to new models requires significant effort, so we only evaluate it on LLAMA-7B, LLAMA2-7B, and GPT-J-6B. 

Table~\ref{tab:compare_pasta} demonstrates that {\tool} consistently outperforms PASTA on five benchmarks, achieving 4X higher Pass@1 on average. 
We attribute this to two hypothetical reasons. 
First, internally editing model attention can be sensitive and exhibit unexpected behaviors.
Second, identifying effective attention headers may not be stable or generalizable across different tasks.
In contrast, {\tool} augments the final logits without internally editing the model's feed-forward process, making it model-agnostic. Furthermore, {\tool} only introduces a single hyperparameter, and the tuning is stable (discussed in Appendix~\ref{app_sub:tuning_stability}).

\section{Confidence-Modulated Anchoring Strength}
\label{app:confidence_modulate}

To make {\tool} more stable and eliminate the need for tuning, we propose a confidence-modulated weighting method that adjusts each token’s anchoring strength based on its original probability respectively.

When the LLM is highly confident about a specific token, the corresponding logit is adjusted less, resulting in a lower anchoring weight applied to the logit difference. Conversely, if the LLM shows lower confidence, a relatively higher weight is applied, leading to a greater adjustment of the logit value. The confidence level is measured using the original probability distribution generated by the LLM.

Formally, let $p_t$ represent the predicted probability of token $t$ within the original probability distribution for the next token. The anchoring strength $\omega_t$ for token $t$ is defined as:
\begin{align}
    \omega_t = \lambda \cdot (1 - p_t),
\end{align}
where $\lambda$ is a fixed coefficient that controls the overall anchoring strength. This results in a token-wise anchoring strength vector $\boldsymbol{\omega} \in \mathbb{R}^{|\mathcal{V}|}$ over the vocabulary $\mathcal{V}$. The vector $\boldsymbol{\omega}$ is then element-wise multiplied with the original logit difference vector to obtain a reweighted logit difference (Figure~\ref{fig:pipeline} \circled{4}).


\end{document}